\documentclass[journal]{IEEEtran}
\newcommand{\ebf}{{\bf e}}

\newcommand{\wbf}{{\bf w}}

\newcommand{\Abf}{{\bf A}}

\newcommand{\Ibf}{{\bf I}}

\newcommand{\Amc}{\mathcal{A}}

\newcommand{\Cmc}{\mathcal{C}}
\newcommand{\Dmc}{\mathcal{D}}
\newcommand{\Emc}{\mathcal{E}}
\newcommand{\Fmc}{\mathcal{F}}
\newcommand{\Gmc}{\mathcal{G}}

\newcommand{\Lmc}{\mathcal{L}}
\newcommand{\Mmc}{\mathcal{M}}
\newcommand{\Nmc}{\mathcal{N}}

\newcommand{\Qmc}{\mathcal{Q}}

\newcommand{\Smc}{\mathcal{S}}

\newcommand{\Umc}{\mathcal{U}}
\newcommand{\Vmc}{\mathcal{V}}

\newcommand{\mubf}{\boldsymbol{\mu}}

\newcommand{\xibf}{\boldsymbol{\xi}}

\newcommand{\varphibf}{\boldsymbol{\varphi}}

\newcommand{\psibf}{\boldsymbol{\psi}}

\newcommand{\Ebb}{\mathbb{E}}
\newcommand{\Rbb}{\mathbb{R}}

\newcommand{\trp}{\intercal}

\newcommand{\NoCons}{M}
\newcommand{\sprime}{p}

\newcommand{\Convex}{1}

\newcommand{\Grad}{3}
\newcommand{\decreaserate}{\zeta}
\IEEEoverridecommandlockouts
\usepackage[dvips]{graphicx}
\usepackage{algorithm}
\usepackage{algorithmic}
\usepackage{psfrag}
\newcommand{\norm}[1]{\left\lVert#1\right\rVert}
\usepackage{bm}
\usepackage{amssymb,mathtools}
\usepackage{amsfonts}
\usepackage{amsmath,amssymb}
\usepackage{amsthm}
\usepackage{cite} 
\usepackage{flushend}
\usepackage[dvipsnames]{xcolor}
\usepackage{url}
\usepackage[hyperfootnotes=false,hidelinks]{hyperref}
\usepackage{graphicx}
\usepackage{upgreek}
\usepackage{yfonts}
\usepackage{cleveref}
\usepackage{caption}
\usepackage{amsmath}
\usepackage{cite}
\usepackage{mathtools}
\usepackage{subfigure}

\begin{document}

\title{Personalized Graph Federated Learning with Differential Privacy}

\author{Francois Gauthier, \textit{Member, IEEE}, Vinay Chakravarthi Gogineni, \textit{Senior Member, IEEE}, Stefan Werner, \textit{Fellow, IEEE}, Yih-Fang Huang, \textit{Life Fellow, IEEE}, Anthony Kuh \textit{Fellow, IEEE}
\thanks{This work was supported by the Research Council of Norway.}
\thanks{A conference precursor of this work appears in the Asilomar conference on signals, systems, and computers, Pacific Grove, USA, Nov. 2022 \cite{Asilomar2}.}
\thanks{F. Gauthier, V. C. Gogineni, and S. Werner are with the Department of Electronic Systems, Norwegian University of Science and Technology, Trondheim, Norway. Email: \{francois.gauthier, vinay.gogineni, stefan.werner\}@ntnu.no.}
\thanks{Yih-Fang Huang is with the Department of Electrical Engineering, University of Notre Dame, Notre Dame, IN 46556 USA (e-mail: huang@nd.edu).}
\thanks{Anthony Kuh is with the Department of Electrical Engineering, University of Hawaii at Manoa, Honolulu, HI 96882 USA (e-mail: kuh@hawaii.edu).}
}

\makeatletter
\newcommand{\linebreakand}{%
  \end{@IEEEauthorhalign}
  \hfill\mbox{}\par
  \mbox{}\hfill\begin{@IEEEauthorhalign}
}
\makeatother

\newcommand{\Lim}[1]{\raisebox{0.5ex}{\scalebox{0.8}{$\displaystyle \lim_{#1}\;$}}}

\maketitle

% Revise plots
% DONE MSE -> NMSD
% DONE E is a set, so null ensemble
% DONE Blue colors to green/red
% DONE Mention ICL and initial privacy value
% DONE Remove extra curve in 2 b
% DONE tau^{(n)}
% ? Redo -> NMSE, check scale, cf above

% Change notations (clean overleaf first)
% DONE C_s^q -> C_{s,(q)} (Give the eq. C_{s,(q)} = C_s \bigcqp C_{(q)})
% DONE g' -> \nabla
% DONE C client no -> K
% DONE Model dim -> L
% DONE not use t -> p
% DONE No of constraints -> M

% DONE Revise paragraphs ending with single word line
% DONE Similarity: Exist between data distribution (or cluster when short), and is exploited between cluster-specific personalized models

% DONE Written following meeting & comments
% DONE Comments details
% DONE Say task similarity instead of data distrib similarity (or modif II A)
% ? Remove punctuation in equations
% Add some more PFL papers
% DONE Rm framework

% DONE New comments most of them
% Comments cf meeting
% MSE cf mail

% Note: It is normal that the impact of ICL does not converge to zero when tau follows a decreasing sequence. The noise absorbed remains.

% Formulation on privacy
% - We quantify the loss (intro, proof, simulations)
% - Phi is the privacy parameter
% - It represents leakage, as said once in the privacy explanation

% G

% Clean plots: (removed plots: R phis)
% - Uniformize axis & legend when poss
% - Axis names
% - Keep same size
% - Captions

% Revise para: copy on Grammarly and take inspiration from Wordtune
%   Try make main subject the first subject

\begin{abstract}
This paper presents a personalized graph federated learning (PGFL) framework in which distributedly connected servers and their respective edge devices collaboratively learn device or cluster-specific models while maintaining the privacy of every individual device. The proposed approach exploits similarities among different models to provide a more relevant experience for each device, even in situations with diverse data distributions and disproportionate datasets. Furthermore, to ensure a secure and efficient approach to collaborative personalized learning, we study a variant of the PGFL implementation that utilizes differential privacy, specifically zero-concentrated differential privacy, where a noise sequence perturbs model exchanges. Our mathematical analysis shows that the proposed privacy-preserving PGFL algorithm converges to the optimal cluster-specific solution for each cluster in linear time. It also shows that exploiting similarities among clusters leads to an alternative output whose distance to the original solution is bounded, and that this bound can be adjusted by modifying the algorithm's hyperparameters. Further, our analysis shows that the algorithm ensures local differential privacy for all clients in terms of zero-concentrated differential privacy. Finally, the performance of the proposed PGFL algorithm is examined by performing numerical experiments in the context of regression and classification using synthetic data and the MNIST dataset.
\end{abstract}

%%%%%%%%%%%%%%%%%%%%%%%%%%%%%%%%%%%%%%%%%%%%%%%%%%%%%%%%%%%%%%%%%%%%%%%%%%%%%%%%

\begin{IEEEkeywords}
Federated learning, personalized learning, graph federated architecture, differential privacy, zero-concentrated differential privacy.
\end{IEEEkeywords}

\section{Introduction}

% Intro
The rise of internet-of-things (IoT) and cyber-physical systems has led to exponential growth in data collection from distributed devices. However, transferring this massive amount of data to a centralized processing point for inference and decision-making is often impractical due to resource constraints and privacy concerns. To overcome these challenges, distributed learning, with its on-device processing, %{\it in situ} processing {\color{red} (forgive me, but I do not know what {\it in situ processing} is)}
is an attractive alternative, enabling efficient data analysis without moving the raw data out of the edge devices. Federated learning (FL) is a distributed learning framework that facilitates collaborative model training across edge devices or clients without exposing the underlying data  \cite{origin, FedOverview, niknam2020federated}. 
In particular, using its local data, each client refines a global model shared by a server, and subsequently transfers the updated model back to the server which then aggregates all updated client models before sending an update back to clients for further refinements.
%Practical implementation of FL faces many challenges that have been studied in the literature, such as privacy \cite{Dwork1, Private, Asilomar}, scalability \cite{SayedGraphFL, Hierarchical, NetworkFL}, communication efficiency \cite{FedAVG, DblCompress, IoTVinay}, asynchronous behavior \cite{CommAsync, AsyncCommSched, ICC}, noisy communication links \cite{Noisy, tuor2021overcoming}, and byzantine attacks \cite{FLByz, FLByzBook}.

% Graph
To date, research on FL mostly uses a single-server architecture, which is susceptible to communication and computation bottlenecks and scales poorly with the number and geographical dispersion of participating clients. To address these concerns, some alternatives to the single-server architecture have been proposed, see, e.g., \cite{SayedGraphFL, Hierarchical, NetworkFL, VinayMultitask}, such as client-edge-server hierarchical learning \cite{Hierarchical} and the graph federated architecture \cite{SayedGraphFL, VinayMultitask}. In client-edge-server hierarchical learning,  edge servers perform partial aggregation with their associated clients and communicate their results to a single cloud server that performs the global aggregation. However, using a single cloud server is susceptible to bottlenecks and can only accommodate up to a limited number of edge servers. In contrast, the graph federated architecture uses a server network in which each server aggregates the information from its associated clients and shares its model with its neighbors. Therefore, the graph federated architecture is highly scalable with the number of clients and easier to implement, thanks to its distributed nature.

% PFL
One of the main challenges in FL is data heterogeneity, which means there can be substantial differences in the underlying statistical distributions among clients' data %Data heterogeneity is one of the most challenging aspects of FL, and in such a large-scale architecture, the underlying data distributions of the clients could exhibit substantial differences 
\cite{FLapplications, AutonomousVehicles, Healthcare}. 
%Furthermore, in such large-scale applications, the participating clients are likely to exhibit differences in their underlying properties and be required to perform device-specific tasks \cite{FLapplications, AutonomousVehicles, Healthcare}. 
Consequently, a unique globally shared model can be inadequate for such settings, and personalized models must be learned instead  %A unique global shared model cannot cater to those differences. Instead, personalized models must be learned 
\cite{PFL, t2020personalized, felix2022personalized}. For example, autonomous vehicles need to maintain vehicle-specific models of their highly dynamic environment while collaborating with nearby vehicles and/or smart city IoT devices \cite{AutonomousVehicles}. This requirement can be met by personalized FL, where clients, or groups of clients (clusters), learn client- or cluster-specific models \cite{IoTMag, PFLmeta, AdaptPFL}. These personalized models typically share some similarities \cite{SayedMT}. As an example, the environment of an autonomous vehicle could be shared with other connected objects. Leveraging the similarities between cluster-specific models can, therefore, improve performance  \cite{SayedMT, VinayTau}, a process known as inter-cluster learning, which is particularly important when some clients or clusters have insufficient data \cite{FLC1s, FLC1s2}. 

Personalized FL has received considerable attention lately due to its ability to improve learning performance in settings where clients are required to observe device-specific behaviors, see, e.g., \cite{SayedMT, FLMT1s, FLMT1s2, OFLMT1s, FLC1s, FLC1s2}. It is used in many applications such as healthcare, electrical load forecasting, biometrics, drone swarms, and autonomous vehicles \cite{Healthcare, ElecLoad, Biometrics, Swarms, AutonomousVehicles}. However, all those works are limited to single-server cases.  For example, although \cite{VinayMultitask} extends personalized FL to a multi-server architecture, it assumes that all the clients associated with a given server learn the same model. Under this assumption, each server maintains a single model trained via conventional FL and the model is refined by communicating with other servers about their models. However, the general case where each distributed server needs to enable the learning of personalized models and collaborate with its neighbors to refine those, is yet to be studied.
%Since neighboring clients may not solve the same learning task, this assumption limits the range of potential applications.

% Privacy
In the context of graph FL, many devices take part in the training process, and ensuring the privacy and security of client data is crucial. The risk of eavesdropping attacks on the client-server channels increases with the number of devices in the system, and not all devices can be trusted. Even if data is not explicitly shared among clients, repeated message exchanges could reveal sensitive information to curious devices or external eavesdroppers \cite{salem2020updates, Private}. In order to reduce this risk, differential privacy (DP) has been introduced to protect client privacy by ensuring that the inclusion or exclusion of an individual data sample does not significantly affect the algorithm output. In other words, DP limits the ability of attackers to infer information about individual data samples by adding controlled noise to the data before sharing it with the server \cite{Dwork1, OverviewDP, Dwork2014, FLPrivacy, DPsequential}. In particular, the zero-concentrated DP (zCDP) variant is well-suited for iterative implementations, as it allows the privacy budget to be adjusted dynamically based on the number of iterations 
%for adds white Gaussian noise with a variance that decreases progressively throughout the computation 
\cite{DworkzCDP, Bun, zCDPsequential, Zhuhan, zCDP_FedAvg, Asilomar}. Therefore, this paper considers zCDP in the graph FL architecture where the privacy of client data is of utmost importance. By employing zCDP, clients perturb their local model estimates  with a noise sequence of known variance that decreases progressively throughout the computation to ensure privacy without compromising model accuracy.

%In a large-scale architecture, not all devices can be trusted, and the numerous communication channels are vulnerable to eavesdroppers. Although data is not shared in FL, curious devices or external eavesdroppers could infer private information from the repeated message exchanges \cite{salem2020updates, Private}. Differential privacy (DP) is designed to protect the participating clients from such attacks by ensuring that the presence or absence of an individual data sample does not impact the output of the algorithm enough for this data sample to be detected. This is achieved by perturbing the local models with noise before sharing them with the server \cite{Dwork1, OverviewDP, Dwork2014, wei2020federated}. Specifically, in the zero-concentrated DP (zCDP) variant, each client perturbs its local estimates with white Gaussian noise, whose variance progressively decreases throughout the computation \cite{DworkzCDP, Bun, Asilomar}. In this paper, we consider the use of zCDP, well-suited for iterative implementations.

% Contribution
This manuscript tackles the general case of personalized graph federated learning (PGFL) in both a conventional and privacy-preserving manner. Specifically, we consider a multi-server architecture with distributed clients grouped into clusters, irrespective of their associated servers, for the decentralized training of cluster-specific personalized models. The proposed algorithms, within the considered PGFL architecture, leverage similarities between clusters to mitigate data scarcity and improve learning performance. The local training in the proposed framework uses the alternating direction method of multipliers (ADMM), well-suited for distributed applications \cite{RecentADMM1, RecentADMM2, ADMMmeta} and demonstrating fast, often linear \cite{shi2014linear, ding2019differentially}, convergence. The main contributions of this manuscript are summarized as follows.
\begin{itemize}
    \item A PGFL framework is proposed to improve learning performance in a distributed learning setting. Our approach employs inter-cluster learning to improve the accuracy of local models by leveraging information from other clusters. The graph FL problem is formulated as a constrained optimization problem and solved in a distributed manner using ADMM.
    %We propose the PGFL algorithm, which uses inter-cluster learning to improve learning performances and the ADMM as the local learning process.
    \item We design a privacy-preserving variant of the PGFL algorithm, where clients perturb their local models to achieve local differential privacy using the zCDP framework. The privacy loss is quantified per iteration as well as throughout the computation.
    %The level of privacy leakage is quantified using the zCDP framework throughout the computation.
    %We implement a privacy-preserving variant of the PGFL, where clients perturb their local models to achieve local differential privacy. The privacy leakage is quantified in the zCDP privacy framework throughout the computation.
    %{\color{red}Mathematical analysis is given to show that the privacy-preserving implementation of the PGFL algorithm can converge to the optimal solution in linear time. Additionally, our analysis shows that the model variation caused by inter-cluster learning is bounded and that the bound depends on cluster similarity and can be adjusted with hyperparameter selection.}
    \item Mathematical analysis is given to show that the privacy-preserving implementation of the PGFL algorithm converges to the optimal solution for each cluster in linear time. Additionally, our analysis shows that utilizing inter-cluster learning leads to an alternative output whose distance to the original solution is bounded and that the bound depends on cluster similarity and can be adjusted with hyperparameter selection.
    %which provides a guarantee of the stability of the algorithm.
    %We carry out mathematical analysis to show that this implementation can converge to the optimal solution in linear time and that the model variation caused by inter-cluster learning is bounded.
\end{itemize}
 The paper is organized as follows. Section II introduces the problem and presents the PGFL algorithm along with its zCDP variant. Sections III and IV are dedicated to the convergence and privacy analyses of the proposed algorithm. In Section V, we demonstrate the effectiveness of the algorithm through a series of experiments involving regression and classification tasks. Section VI concludes the paper.
 
% Math section

\textit{Mathematical notations:} Matrices, column vectors, and scalars are denoted by bold uppercase, bold lowercase, and lowercase letters, respectively. The notation $\Abf^\mathsf{T}$ denotes transpose of the matrix $\Abf$, the identity matrix is denoted by ${\bf I}$, and a null vector by ${\bf 0}$. The exclusion of an element $a$ from set $\Amc$ is denoted $\Amc \backslash a$. The notation $\langle \boldsymbol{a},\boldsymbol{b} \rangle$ denotes the inner product between vectors $\boldsymbol{a}$ and $\boldsymbol{b}$. The statistical expectation operator is represented by $\Ebb[\cdot]$, and $\Nmc(\boldsymbol{\mu},\boldsymbol{\Sigma})$ and $\Umc(a,b)$ respectively denote the normal distribution with mean $\boldsymbol{\mu}$ and covariance matrix $\boldsymbol{\Sigma}$ and the uniform distribution on an interval $(a,b)$. Finally, the gradient of a function $a(\cdot)$ is denoted by $\nabla a(\cdot)$.
% so that $\Qmc \backslash q = \Qmc - \{ q \}$
%If a random variable $A$ follows the law $\mathcal{B}$, we will write $A \sim \mathcal{B}$

%%%%%%%%%%%%%%%%%%%%%%%%%%%%%%%%%%%%%%%%%%%%%%%%%%%%%%%%%%%%%%%%%%%%%%%%%%%%%%%%
\section{Problem Formulation and Proposed Method}
The proposed PGFL framework solves a personalized optimization problem in a graph federated architecture and utilizes the similarities among clusters to enhance learning performance. For this purpose, we consider a distributed network of $S$ servers associated with a total of $K$ clients.
The server network is modeled as an undirected graph $\Gmc = (\Smc, \Emc)$, where $\Smc$ is the set of servers and $\Emc$ is the set of edges so that two servers $s$ and $\sprime$ can communicate if and only if $(s, \sprime) \in \Emc$.
The set of neighbors to a server $s$ is denoted $\Nmc_s$, and contains $s$, we denote $\Nmc_s^- = \Nmc_s \backslash s$.
Each server $s$ is associated with a set of clients, denoted $\Cmc_s$, with $\bigcup_{s \in \Smc} \Cmc_s = \Cmc$ and $\Cmc_s \bigcap \Cmc_\sprime = \emptyset, \forall s \neq \sprime $. Every client $k \in \Cmc$ has access to a local dataset $\mathcal{D}_k$ of cardinality $|\mathcal{D}_k| = D_k$, which is composed of a data matrix ${\bf X}_k = [{\bf x}_{k,1} \hdots {\bf x}_{k,D_k}]^\trp$, where ${\bf x}_{k,i}, i \in \{1, \hdots, D_k \}$ is a vector of size $L$, and a response vector ${\bf y}_k = [y_{k,1}, \hdots, y_{k,D_k}]^\trp$ that is subject to white observation noise. Each client $k \in \Cmc$ aims to learn a personalized, client-specific model $\wbf_{k}$.

% New edit
%{\color{red}The learning task for each client is defined by the set $\{ \Dmc_k, \ell_k, R_k \}$, which represents its local data, loss function, and regularizer function.}
The learning task for each client is defined by the set $\{ \Dmc_k, \ell_k \}$, which represents its local data and loss function. All clients connected to distributed servers, regardless of their associated servers, are grouped into $Q$ clusters. These clusters are formed by clients with similar learning tasks, such as $\Fmc$-similar tasks \cite{ben2003exploiting}, with the aim of collectively learning a shared model. It is assumed that there is a degree of relationship among the learning tasks across clusters, which can manifest in various ways. For example, clusters may share the same loss and regularizer functions while having different data distributions, or they may have the same data distribution but distinct objective functions. For instance, in healthcare, clusters can represent various patient diagnostics, independent of their respective associated hospitals, with a hospital functioning akin to a server. We denote the set of clusters as $\Qmc = \{ 1, \hdots, Q \}$. The clients belonging to a specific cluster $q \in \Qmc$  form the set $\Cmc_{(q)}$ aiming to learn the model $\wbf_q^*$. Additionally, the set of clients associated with server $s$ within cluster $q$ is denoted as $\Cmc_{s,(q)}$, with $\Cmc_{s,(q)} = \Cmc_{s} \bigcap \Cmc_{(q)}$.

\subsection{Personalized Graph Federated Learning}
To address task variations, personalized (cluster) models are preferable. However, despite these differences, the underlying relationship among tasks, or equivalently, clusters, can still be exploited in decentralized learning. Here, we consider a modified regularized empirical risk minimization problem to leverage similarities among the clusters. For this purpose, we introduce an additional regularizer function that enforces similarity among the cluster-specific personalized models. This additional regularizer function corresponds to inter-cluster learning and is controlled by the inter-cluster learning parameter $\tau \in (0, 1)$. The resulting optimization problem for a cluster $q$ is formulated as:
\begin{align}
    \label{ERM_global}
    \min_{\wbf_q} &\sum_{k \in \Cmc_{(q)}} \frac{1}{D_k} \sum_{i=1}^{D_k} \ell_k({\bf x}_{k,i}, y_{k,i}; \wbf_q) + \lambda R(\wbf_q) \notag \\
    &+ \tau \sum_{r \in \Qmc \backslash q} || \wbf_{r} - \wbf_q ||_2^2 ,
\end{align}
where $\ell_k(\cdot)$, $R(\cdot)$, and $\lambda$ denote the client loss function, the global regularizer function, and the regularization parameter, respectively. The larger the $\tau$ value is, the more the similarities among cluster-specific personalized models are exploited.

%{\color{red}reach consensus by communicating with their neighbors.}
The centralized optimization problem above relies on the global variable $\wbf_q$. In a multi-server architecture, the servers maintain local cluster-specific models and communicate among neighbors to reach a consensus for each cluster. The equivalent distributed optimization problem for a server $s$ and cluster $q$ is
\begin{align}
    \label{ERM_servers}
    &\underset{\wbf_{q,s}}{\min}
    && \sum_{k \in \Cmc_{s,(q)}} \frac{1}{D_k} \sum_{i=1}^{D_k} \ell_k({\bf x}_{k,i}, y_{k,i}; \wbf_{q,s}) + \lambda R(\wbf_{q,s}) \notag \\
    & 
    && \hspace{8mm} + \tau \sum_{r \in \Qmc \backslash q} \sum_{\sprime \in \Nmc_s} || \wbf_{r,\sprime} - \wbf_{q,s} ||_2^2, \\
    &\text{\ s.t.}\ 
    && \wbf_{q,s} = {\bf z}_{s,\sprime}, \wbf_{q, \sprime} = {\bf z}_{s,\sprime}; \forall (s, \sprime) \in \Emc, \notag
\end{align}
where $\wbf_{q,s}$ denotes the model of server $s$ for cluster $q$ and consensus is enforced by the auxiliary variables $\{ {\bf z}_{s,\sprime}; \forall (s, \sprime) \in \Emc \}$. From \eqref{ERM_servers}, the augmented Lagrangian with penalty parameter $\rho$ can be derived as
\begin{align}
    \label{Lagrangian}
    \Lmc_{\rho,q} &(\Vmc_q, \mathcal{M}, \mathcal{Z}) = \sum_{s \in \Smc} \Biggl[ \sum_{k \in \Cmc_{s,(q)}} \frac{\ell_k({\bf X}_{k}, {\bf y}_{k}; \wbf_{q,s})}{D_k} + \lambda R(\wbf_{q,s}) \notag \\
    & + \tau \sum_{r \in \Qmc \backslash q} \sum_{\sprime \in \Nmc_s} || \wbf_{r,\sprime} - \wbf_{q,s} ||_2^2 \notag \\
    &+ \sum_{\sprime \in \Nmc_s^-} \Bigl( \mubf_{s,\sprime}^\trp (\wbf_{q,s} - {\bf z}_{s,\sprime}) + \psibf_{s,\sprime}^\trp (\wbf_{q, \sprime} - {\bf z}_{s,\sprime}) \Bigr) \notag \\
    &+ \frac{\rho}{2} \sum_{\sprime \in \Nmc_s^-} \Bigl( || \wbf_{q,s} - {\bf z}_{s,\sprime} ||_2^2 + || \wbf_{q, \sprime} - {\bf z}_{s,\sprime} ||_2^2 \Bigr) \Biggr],
\end{align}
with the set of primal variables $\Vmc_q = \{ \wbf_{q,s}; s \in \Smc \}$, Lagrange multipliers $\mathcal{M} = ( \{ \mubf_{s,\sprime} \}, \{ \psibf_{s,\sprime}\}) $, and auxiliary variables $ \mathcal{Z} = \{ {\bf z}_{s,\sprime} \}$. Given that the Lagrange multipliers are initialized to zero, using the Karush-Kuhn-Tucker conditions of optimality and setting $\psibf_s = 2 \sum_{\sprime \in \Nmc_s^-} \psibf_{s,\sprime}$, it can be shown that the Lagrange multipliers $\mubf_{s,\sprime}$ and the auxiliary variables $\mathcal{Z}$ are eliminated \cite{Giannakis2016}. From \eqref{Lagrangian}, it is possible to derive the local update steps of the ADMM for clients and servers. For client $k \in \Cmc_{s,(q)}$, the primal and dual updates are given by
\begin{itemize}
    \item[] \textbf{Client primal update:}
\begin{align}
    \label{ClientPrimalUpdate}
    \wbf_k^{(n)} = &\text{arg} \min_{\wbf} \frac{1}{D_k} \ell_k({\bf X}_{k}, {\bf y}_{k}; \wbf) + \frac{\lambda}{|\Cmc_s|} R(\wbf) \\
    &- \left< \varphibf_k^{(n-1)}, \wbf - \wbf_{q,s}^{(n-1)} \right> + \frac{\rho}{2} ||\wbf - \wbf_{q,s}^{(n-1)}||_2^2, \notag
\end{align}
    \item[] \textbf{Client dual update:}
\begin{align}
    \label{ClientDualUpdate}
    \varphibf_k^{(n)} = \varphibf_k^{(n-1)} + \rho ( \wbf_{q,s}^{(n)} - \wbf_k^{(n)} ),
\end{align}
\end{itemize}
where the superscript $n$ denotes the iteration number. Further, the primal and dual updates for a server $s \in \Smc$ are given by:

\begin{algorithm}[t!]
\caption{PGFL}
\label{PGFL}
\begin{algorithmic}[1]
    \item[] \textbf{Initialization:} $\wbf_k^{(0)} = {\bf 0}$ and $\wbf_{q,s}^{(0)} = 0, \forall k, q, s$
    \item[] \textit{-- Procedure at client $k \in \Cmc_{s}$ --} 
    \item[] \textbf{For} iteration $n=1,2,\hdots$
        \item[] \hspace{5mm} Update $\wbf_k^{(n)}$ as in \eqref{ClientPrimalUpdate}
        \item[] \hspace{5mm} Share $\wbf_k^{(n)}$ and $\varphibf_k^{(n-1)}$ with server $s$
        \item[] \hspace{5mm} Receive $\wbf_{q,s}^{(n)}$ from server $s$
        \item[] \hspace{5mm} Update $\varphibf_k^{(n)}$ as in \eqref{ClientDualUpdate}
    \item[] \textbf{EndFor}
    \item[] \textit{-- Procedure at server $s$ --} 
    \item[] \textbf{For} iteration $n=1,2,\hdots$
        \item[] \hspace{5mm} Receive $\{ \widetilde{\wbf}_k^{(n)}, \varphibf_k^{(n-1)}; \forall k \in \Cmc_s \}$
        \item[] \hspace{5mm} Update $\widetilde{\wbf}_{q,s}^{(n)}$ as in \eqref{ServerAggregation}
        \item[] \hspace{5mm} Share $\widetilde{\wbf}_{q,s}^{(n)}, \forall q$ with each server $\sprime$ in $\Nmc_s^-$
        \item[] \hspace{5mm} Receive $\widetilde{\wbf}_{q, \sprime}^{(n)}, \forall q$ from each server $\sprime$ in $\Nmc_s^-$
        \item[] \hspace{5mm} Aggregate $\widehat{\wbf}_{q,s}^{(n)}$ as in \eqref{InterServerAggregation}
        \item[] \hspace{5mm} Compute $\wbf_{q,s}^{(n)}$ as in \eqref{inter-clusterLearning}
        \item[] \hspace{5mm} Share $\wbf_{q,s}^{(n)}$ with clients in $\Cmc_s$
    \item[] \textbf{EndFor}
\end{algorithmic} 
\end{algorithm}
%\begin{itemize}
%   \item[] \textbf{Server primal update:}
\begin{align}
    \label{ServerPrimal}
    % Aggregation & Dual impact (inter-server homogeneity)
    %\wbf_{q,s}^{(n)} &= \text{arg} \min_{\wbf} \Biggl[ \norm{ \wbf - \widetilde{\wbf}_{q,s}^{(n)} }_2^2 + \wbf^\trp \psibf_{q,s}^{(n-1)} \notag\\ \notag \\
    % Inter cluster learning (can be seen as regularizer)
    %&\: + \tau^{(n)} \Bigl| \Bigl| \wbf - \frac{1}{Q-1} \frac{1}{|\Nmc_s|} \sum_{r \in \Qmc \backslash q}  \sum_{\sprime \in \Nmc_s}\wbf_{r,\sprime}^{(n-1)} \Bigl| \Bigl|_2^2 \notag \\
    % Augmented lagrangian effect
    %&\: + \rho \sum_{\sprime \in \Nmc_s^-} \norm{ \wbf - \frac{\wbf_{q,s}^{(n-1)} - \wbf_{q, \sprime}^{(n-1)}}{2} }_2^2 \Biggr], \\
% Solution
    % Aggregation
    \wbf_{q,s}^{(n)} &= \frac{1}{1 + \tau^{(n)} + \rho |\Nmc_s^-|} \Biggl[ \frac{1}{|\Cmc_{s,(q)}|} \sum_{k \in \Cmc_{s,(q)}} \wbf_k^{(n)} \notag \\
    &- \frac{1}{\rho |\Cmc_{s,(q)}|} \sum_{k \in \Cmc_{s,(q)}} \varphibf_k^{(n-1)} \notag \\
    % Inter-server aggregation with dual
    &- \frac{1}{2} \psibf_{q,s}^{(n-1)} + \frac{\rho}{2} \sum_{\sprime \in \Nmc_s^-} ( \wbf_{q,s}^{(n-1)} - \wbf_{q, \sprime}^{(n-1)} ) \notag \\
    % Inter cluster learning (can be seen as regularizer)
    &+ \tau^{(n)} \frac{1}{Q-1} \frac{1}{|\Nmc_s|} \sum_{r \in \Qmc \backslash q}  \sum_{\sprime \in \Nmc_s}\wbf_{r,\sprime}^{(n-1)} \Biggr], \\
%\end{align}
%\item \textbf{Server dual update}
%\begin{align}
    \psibf_{q,s}^{(n)} &= \psibf_{q,s}^{(n-1)} + \rho \sum_{\sprime \in \Nmc_s^-} \Bigl( \wbf_{q, \sprime}^{(n)} - \wbf_{q,s}^{(n)} \Bigr),
\end{align}
%\end{itemize}
%with $\widetilde{\wbf}_{q,s}^k$ given by
%\begin{align}
%    \widetilde{\wbf}_{q,s}^{(n)} =  \frac{1}{|\Cmc_{s,(q)}|} \sum_{k \in \Cmc_{s,(q)}} \wbf_k^{(n)} - \frac{1}{\rho |\Cmc_{s,(q)}|} \sum_{k \in \Cmc_{s,(q)}} \varphibf_k^{(n-1)}.
%\end{align}
where $\tau^{(n)}$, the inter-cluster learning parameter, is iteration-dependent. Since inter-cluster learning may degrade performance toward the end of the computation, it may be necessary for $\tau^{(n)}$ to follow a decreasing sequence.

%As inter-cluster learning may degrade performances towards the end of the computation, this inter-cluster learning parameter will be time-varying.
%The inter-cluster learning parameter $\tau^{(n)}$ is iteration-dependent since it may be necessary for it to follow a decreasing sequence to avoid over-leveraging the task similarities.

The computation in \eqref{ServerPrimal} performs local aggregation (first two lines), inter-server aggregation (third line), and inter-cluster learning (fourth line) in a single step. This presents the major drawback of using the models of the previous iteration for inter-server aggregation, i.e., $\wbf_{q, \sprime}^{(n-1)}$, and inter-cluster learning, i.e., $\wbf_{r, \sprime}^{(n-1)}$ \cite{StaticTauSayed, StaticTauVinay}. A multi-step mechanism addresses this issue by replacing the primal and dual updates of the server as follows:
\begin{itemize}
    \item \textbf{Server aggregation}
\begin{align}
    \label{ServerAggregation}
    \widetilde{\wbf}_{q,s}^{(n)} = \frac{1}{|\Cmc_{s,(q)}|} \sum_{k \in \Cmc_{s,(q)}} \wbf_k^{(n)} - \frac{1}{\rho |\Cmc_{s,(q)}|} \sum_{k \in \Cmc_{s,(q)}} \varphibf_k^{(n-1)}.
\end{align}
\item \textbf{Inter-server aggregation}
\begin{align}
    \label{InterServerAggregation}
    \widehat{\wbf}_{q,s}^{(n)} = \frac{1}{|\Nmc_s|} \sum_{\sprime \in \Nmc_s} \widetilde{\wbf}_{q, \sprime}^{(n)}.
\end{align}
\item \textbf{Inter-cluster learning}
\begin{align}
    \label{inter-clusterLearning}
    &\wbf_{q,s}^{(n)} = \Bigl( 1 - \tau^{(n)} \Bigr) \widehat{\wbf}_{q,s}^{(n)} + \frac{\tau^{(n)}}{Q-1} \sum_{r \in \Qmc \backslash q} \widehat{\wbf}_{r,s}^{(n)}.
\end{align}
\end{itemize}

The above multi-step mechanism has two main advantages. First, performing server aggregation prior to inter-server aggregation enables the servers to maintain models composed of the last available client estimates. Second, the fact that inter-cluster learning is performed at the end of the multi-step mechanism ensures that model similarities are leveraged evenly; that is, the same weight is given to any two clients' estimates within the server neighborhood. The resulting PGFL algorithm is summarized in Algorithm \ref{PGFL}. %This algorithm presents the computation and communication steps that clients and servers perform at each iteration.

%%%%%%%%%%%%%%%%%%%%%%%%%%%%%%
\subsection{Privacy Preservation in PGFL}

We propose a privacy-preserving variant of the PGFL algorithm that implements zero-concentrated differential privacy (zCDP). The motivation for using zCDP, as opposed to the conventional $(\epsilon, \delta)$-DP, is that it provides better accuracy for identical privacy loss under the worst-case scenario that an eavesdropper aggregates all the exchanged messages \cite{Dwork, DworkzCDP}. Instead of sharing the exact local estimate $\wbf_k^{(n)}$, a client $k$ shares with its server the perturbed estimate $\widetilde{\wbf}_k^{(n)}$, given by
\begin{align}
    \label{PrivacyNoiseAddition}
    \widetilde{\wbf}_k^{(n)} = \wbf_k^{(n)} + \xibf_k^{(n)},
\end{align}
where the perturbation noise follows a Gaussian mechanism, $\xibf_k^{(n)} \sim \Nmc(\mathbf{0}, \delta_k^{2(n)} \Ibf)$, with $\delta_k^{2(n)}$ being the variance of the perturbation noise at iteration $n$.

\begin{algorithm}[t!]
\caption{Privacy-preserving PGFL}
\label{PPGFL}
\begin{algorithmic}[1]
    \item[] \textbf{Initialization:} $\wbf_k^{(0)} = {\bf 0}$ and $\wbf_{q,s}^{(0)} = {\bf 0}, \forall k, q, s$
    \item[] \textit{-- Procedure at client $k \in \Cmc_s$ --} 
    \item[] \textbf{For} iteration $n=1,2,\hdots$
        \item[] \hspace{5mm} Update $\wbf_k^{(n)}$ as in \eqref{ClientPrimalUpdate}
        \item[] \hspace{5mm} Perturb $\wbf_k^{(n)}$ into $\widetilde{\wbf}_k^{(n)}$ as in \eqref{PrivacyNoiseAddition}
        \item[] \hspace{5mm} Share $\widetilde{\wbf}_k^{(n)}$ and $\varphibf_k^{(n-1)}$ with server $s$
        \item[] \hspace{5mm} Receive $\wbf_{q,s}^{(n)}$ from server $s$
        \item[] \hspace{5mm} Update $\varphibf_k^{(n)}$ as in \eqref{ClientDualUpdate} using $\widetilde{\wbf}_{q,s}^{(n)}$ and $\widetilde{\wbf}_k^{(n)}$.
        %\eqref{PrivateClientDualUpdate}
    \item[] \textbf{EndFor}
    \item[] \textit{-- Procedure at server $s$ --} 
    \item[] \textbf{For} iteration $n=1,2,\hdots$
        \item[] \hspace{5mm} Receive $\{ \widetilde{\wbf}_k^{(n)}, \varphibf_k^{(n-1)}; \forall k \in \Cmc_s \}$
        \item[] \hspace{5mm} Update $\widetilde{\wbf}_{q,s}^{(n)}$ as in \eqref{PrivateServerAggregation}
        \item[] \hspace{5mm} Share $\widetilde{\wbf}_{q,s}^{(n)}, \forall q$ with each server $\sprime$ in $\Nmc_s^-$ %$\{\sprime, \forall \sprime \in \Nmc_s^- \}$
        \item[] \hspace{5mm} Receive $\widetilde{\wbf}_{q, \sprime}^{(n)}, \forall q$ from each server $\sprime$ in $\Nmc_s^-$
        \item[] \hspace{5mm} Aggregate $\widehat{\wbf}_{q,s}^{(n)}$ as in \eqref{InterServerAggregation}
        \item[] \hspace{5mm} Compute $\wbf_{q,s}^{(n)}$ as in \eqref{inter-clusterLearning}
        \item[] \hspace{5mm} Share $\wbf_{q,s}^{(n)}$ with clients in $\Cmc_s$
    \item[] \textbf{EndFor}
\end{algorithmic} 
\end{algorithm}

%Several different strategies may be considered in the choice of the value of $\delta_k^{2(n)}$. Different strategies are better suited to different types of attacks and to different algorithm complexity. 
%Regardless of the chosen perturbation strategy, aggregating several messages enables an eavesdropper to infer more information. In conventional DP, titled $(\epsilon, \delta)$-DP for its two parameters $\epsilon$ and $\delta$, the privacy protection does not decrease throughout the computation. The privacy loss comes from the potential aggregation of messages \cite{Dwork}. In zCDP, noise variance $\delta_k^{2(n)}$ decreases linearly with the number of iterations; about as much information can be extracted from the last estimate as from the aggregated messages. Therefore, under the assumption that the eavesdropper aggregates all the available messages, zCDP allows for better accuracy than $(\epsilon, \delta)$-DP without privacy loss. R\'enyi-DP is a relaxation of zCDP which concentrates on a single moment of a privacy loss variable \cite{RenyiDP}, as opposed to zCDP, which provides a linear bound on all positive moments. Its purpose is to facilitate analysis in complex machine learning applications, which is unnecessary for this work.

%We assume the worst-case scenario regarding potential eavesdroppers, that is, that they can aggregate several messages to perform reconstruction attacks. For this reason, we use zCDP, 

In the context of zCDP, privacy protection is governed by $\phi_k^{(0)}$ and $\decreaserate$. The parameter $\phi_k^{(0)}$ represents the initial privacy leakage, indicating the desired level of privacy at the start of the algorithm. On the other hand, $\decreaserate \in (0,1)$ denotes the exponential decay factor of the noise variance, determining how the privacy budget diminishes over successive iterations. As shown later in Section IV, the privacy parameter at iteration $n$, $\phi_k^{(n)}$, is inversely proportional to the variance of the perturbation noise, $\delta_k^{2(n)}$. In other words, the privacy parameter decreases as the noise variance increases, providing a stronger privacy guarantee. Conversely, the privacy parameter increases as the noise variance decreases, implying a weaker privacy guarantee. Here, for each client, $k \in \Cmc$, the initial variance $\delta_{k}^{2(0)}$ is fixed, and subsequently, the variance at iteration $n$ is updated according to the relationship $\delta_k^{2(n)} = \decreaserate \delta_{k}^{2(n-1)}$. This recursive update ensures a decreasing privacy budget as the algorithm progresses.
%In zCDP, privacy protection is controlled by two parameters, $\phi_k^{(0)}$ and $\decreaserate$. Where $\phi_k^{(0)}$ is the initial privacy value and $\decreaserate \in (0,1)$ is the exponential decay factor of the variance. For any client $k \in \Cmc$, $\delta_{k}^{2(0)}$ is fixed and $\delta_k^{2(n)} = \decreaserate \delta_{k}^{2(n-1)}$. 

The server aggregation \eqref{ServerAggregation} and client dual update \eqref{ClientDualUpdate} are affected by the noise perturbation \eqref{PrivacyNoiseAddition}. The server aggregation becomes
\begin{align}
    \label{PrivateServerAggregation}
    \widetilde{\wbf}_{q,s}^{(n)} = \frac{1}{|\Cmc_{s,(q)}|} \sum_{k \in \Cmc_{s,(q)}} \widetilde{\wbf}_k^{(n)} - \frac{1}{\rho |\Cmc_{s,(q)}|} \sum_{k \in \Cmc_{s,(q)}} \varphibf_k^{(n-1)},
\end{align}
and in the client dual update, we substitute $\wbf_{q,s}^{(n)}$ with $\widetilde{\wbf}_{q,s}^{(n)}$ and $\wbf_k^{(n)}$ with $\widetilde{\wbf}_k^{(n)}$.
%{\color{red} (I changed from "the decrease rate of noise variance" to "the rate of decreasing noise variance", but I wonder if my change altered what $\decreaserate$ really means.  Also, $\decreaserate$ has not been previously defined.  Where does it fit in the equation (12)?) {\color{green} (Stefan's comment): Francois, I understand from the text that the variance is exponentially decaying and where $\decreaserate$ is the "decay factor. The explanation is a bit convoluted, especially if this is only a special case."}}
%\begin{itemize}
%    \item \textbf{Server aggregation}
%\begin{align}
%    \label{PrivateServerAggregation}
%    \widetilde{\wbf}_{q,s}^{(n)} = \frac{1}{|\Cmc_{s,(q)}|} \sum_{k \in \Cmc_{s,(q)}} \widetilde{\wbf}_k^{(n)} - \frac{1}{\rho |\Cmc_{s,(q)}|} \sum_{k \in \Cmc_{s,(q)}} \varphibf_k^{(n-1)},
%\end{align}
%    \item \textbf{Client dual update}
%\begin{align}
%    \label{PrivateClientDualUpdate}
%    \varphibf_k^{(n)} = \varphibf_k^{(n-1)} + \rho \Bigl( \wbf_{q,s}^{(n)} - \widetilde{\wbf}_k^{(n)} \Bigr).
%\end{align}
%\end{itemize}

The resulting privacy-preserving algorithm is summarized in Algorithm. \ref{PPGFL}. In the following sections, we provide a detailed study of the privacy protection and convergence properties of the proposed privacy-preserving PGFL algorithm.

%%%%%%%%%%%%%%%%%%%%%%%%%%%%%%%%%%%%%%%%%%%%%%%%%%%%%%%%%%%%%%%%%%%%%%%%%%%%%%%%
\section{Convergence Analysis}

%{\color{red}bounds the variation of the cluster-specific personalized models caused by inter-cluster learning by a function of the inter-cluster learning parameter sequence.}
This section studies the convergence behavior of the proposed privacy-preserving PGFL algorithm. Sections III-A and III-B study the algorithm without inter-cluster learning and show that it converges to the optimal solution of \eqref{ERM_servers} with $\tau = 0$ in linear time. Section III-C bounds the distance between the cluster-specific solutions obtained with and without inter-cluster learning by a function of the inter-cluster learning parameter sequence.
%therefore bounding the distance to the optimal solution when using inter-cluster learning.
%Section III-C bounds the difference between the cluster-specific personalized models with and without inter-cluster learning, therefore bounding the distance to the optimal solution when using inter-cluster learning.
%Section III-C bounds the impact of inter-cluster learning on the resulting cluster-specific models, therefore bounding the distance to the optimal solution when leveraging similarities between clusters to improve learning speed.

%%%%%%%%%%%%%%%%%%%%%%%%%%%%%%%%%%%%%%%%
\subsection{Problem Reformulation} % If remove check for section III C in simulations

We consider the server update steps with $\tau^{(n)} = 0$. Then, the minimization problem solved at a client $k \in \Cmc_{s,(q)}$ becomes
\begin{align}
    \label{ClientOpti}
    &\min_{\wbf_k} \frac{1}{D_k} \ell_k({\bf X}_{k}, {\bf y}_{k}; \wbf_k) + \frac{\lambda}{|\Cmc_s|} R(\wbf_k) \notag \\
    &\text{s.t. } \wbf_k = \widehat{\wbf}_{q,s},
\end{align}
where $\widehat{\wbf}_{q,s}$ is the result of inter-server aggregation \eqref{InterServerAggregation}, defined as the average model for cluster $q$ in $\Nmc_s$. To simplify the analysis, we reformulate \eqref{ClientOpti} as
%using the auxiliary variables $\{ \ebf_{k,l} \},  \forall k,l \in \sum_{\sprime \in \Nmc_s} \Cmc_{\sprime, (q)}$, yielding the equivalent formulation:

%Enforcing $\wbf_k = \widehat{\wbf}_{q,s}, \forall k \in \Cmc_{s,(q)}$ has the same effect as enforcing equality between each couple of clients present in the aggregation using the intermediary variables $\{ \ebf_{k,l} \},  \forall k,l \in \sum_{\sprime \in \Nmc_s} \Cmc_{\sprime, (q)}$. We note that the model evolution might be slightly different using this method, so we take the following assumption to simplify the analysis.
% Can motivate further saying
% - Lagrangian use L2 which will yield average
% - This is actually better
% Talk about the dual ?

%\vspace{2pt}
%\noindent \textbf{Assumption \Consensus. (Consensus constraints)} \textit{For a given cluster $q$ and server $s$, the constraints $\wbf_k = \widehat{\wbf}_{q,s}, \forall k \in \Cmc_{s,(q)}$ are equivalent to $\wbf_k = \ebf_{k,l}, \wbf_l = \ebf_{k,l}, \forall k,l \in \sum_{\sprime \in \Nmc_s} \Cmc_{\sprime, (q)}$.}
%\vspace{2pt}

%Under Assumption \Consensus, \eqref{ClientOpti} can be reformulated as
\begin{align}
    \label{ClientOptiWConstraints}
    &\min_{\wbf_k} f_k(\wbf_{k}) \notag \\
    &\text{s.t. } \wbf_k = \ebf_{k,l}, \wbf_l = \ebf_{k,l}, \forall l \in \sum_{\sprime \in \Nmc_s} \Cmc_{\sprime, (q)},
\end{align}
where $f_k(\wbf_{k})$ is given by
\begin{align}
    &f_k(\wbf_{k}) = \frac{1}{D_k} \ell_k({\bf X}_{k}, {\bf y}_{k}; \wbf_k) + \frac{\lambda}{|\Cmc_s|} R(\wbf_{k}),
\end{align}
and the auxiliary variables $\{ \ebf_{k,l} \},  \forall k,l \in \sum_{\sprime \in \Nmc_s} \Cmc_{\sprime, (q)}$ enforce consensus. To reformulate \eqref{ClientOptiWConstraints} further, we introduce the following:
\begin{align}
    \label{Client_Reformulation}
    &\wbf = [\wbf_{1}^\trp, \hdots, \wbf_{k}^\trp, \hdots \wbf_{|\Cmc|}^\trp]^\trp, \notag \\
    &\widetilde{\wbf} = [\widetilde{\wbf}_{1}^\trp, \hdots, \widetilde{\wbf}_{k}^\trp, \hdots \widetilde{\wbf}_{|\Cmc|}^\trp]^\trp \notag  = \wbf + \boldsymbol{\xi} \notag \\
    &\varphibf = [\varphibf_{1}^\trp, \hdots, \varphibf_{k}^\trp, \hdots, \varphibf_{|\Cmc|}^\trp]^\trp, \notag \\
    &F(\wbf) = \sum_{k \in \Cmc} f_k(\wbf_{k}),
\end{align}
where $\boldsymbol{\xi}$ is the concatenation of the noise added to the local models to ensure privacy. In addition, we introduce the vector $\ebf \in \mathbb{R}^{2 \NoCons d}$ concatenating the vectors $\ebf_{k,l}, \ebf_{l,k}, \forall (k,l) \in \{1, \hdots, K \} : k \neq l$, where $d$ is the dimension of the models and $\NoCons$ is the number of constraints in \eqref{ClientOptiWConstraints}. We can then reformulate \eqref{ClientOptiWConstraints} as
% twice over (that is, $\ebf_{k,l}$ and $\ebf_{l,k}$ are distinct)
\begin{align}
    \label{ReformulatedClientOpti}
    &\min_{\wbf} F(\wbf) \notag \\
    &\text{s.t. } {\bf A} \wbf + {\bf B} \ebf = {\bf 0}.
\end{align}
where ${\bf A} = [{\bf A}_1, {\bf A}_2]$ and ${\bf B} = [- {\bf I}_{2 \NoCons d}, - {\bf I}_{2 \NoCons d}]$. The matrices ${\bf A}_1, {\bf A}_2 \in \Rbb^{2 \NoCons d \times |\Cmc| d}$ are composed of $d \times d$-sized blocks. Given a couple of connected clients $(k, l)$, their associated auxiliary variable $\ebf_{k,l}$, and its corresponding index in $\ebf$, $q$; the blocks $ \bigl( {\bf A}_1 \bigr)_{q, k} $ and $ \bigl( {\bf A}_2 \bigr)_{q, l} $ are equal to the identity matrix ${\bf I}_d$, all other blocks are null.

%matrices where if $\ebf_{k,l}$ is the $q^{\text{th}}$ block of $\ebf$, then block $(q, k)$ in ${\bf A}_1$ and block $(q, l)$ in ${\bf A}_2$ are equal to the identity matrix ${\bf I}_d$; all other blocks are null. Finally, the matrix ${\bf B}$ is given by .

%We introduce matrices  ${\bf A}_1, {\bf A}_2 \in \mathcal{M}^{2 \NoCons L \times |\Cmc| d}$ that are $d \times d$-sized block matrices where if $\ebf_{k,l}$ is the $q^{\text{th}}$ block of $\ebf$, then block $(q, k)$ in ${\bf A}_1$ and block $(q, l)$ in ${\bf A}_2$ are equal to the identity matrix ${\bf I}_d$; all other blocks are null. Further, we introduce the matrices ${\bf A} = [{\bf A}_1, {\bf A}_2]$ and ${\bf B} = [- {\bf I}_{2 \NoCons d}, - {\bf I}_{2 \NoCons d}]$ so that the optimization problem for the whole network can be expressed as

From the above definitions, one can express $\sum_{\ebf_{k,l} \in \ebf} \norm{\wbf_k - \ebf_{k,l}}^2 + \norm{\wbf_l - \ebf_{k,l}}^2 = \norm{{\bf A} \wbf + {\bf B} \ebf}^2$ and, for $\boldsymbol{\lambda} \in \mathbb{R}^{4 \NoCons d}$, $\sum_{k \in \Cmc} \sum_{l \in \Nmc_k} (\langle \wbf_k - \ebf_{k,l}, \boldsymbol{\lambda}_q \rangle + \langle \wbf_l - \ebf_{k,l}, \boldsymbol{\lambda}_{2 e + q} \rangle) = \langle {\bf A} \wbf + {\bf B} \ebf, \boldsymbol{\lambda} \rangle$. 

Therefore, the Lagrangian can be rewritten as 
\begin{align}
    \label{LagrangianReformulated}
    \Lmc_{\rho} (\Vmc_q, \Mmc) = F(\wbf) + \langle {\bf A} \wbf + {\bf B} \ebf, \boldsymbol{\lambda} \rangle + \frac{\rho}{2} \norm{{\bf A} \wbf + {\bf B} \ebf}^2.
\end{align}

%%%%%%%%%%%%%%%%%%%%%%%%%%%%%%%%%%%%%%%%
\subsection{Convergence Proof}

\noindent We make the following assumptions to continue the analysis.

\vspace{2pt}
\noindent \textbf{Assumption \Convex.} \textit{The functions $f_{k}(\cdot), k \in \{1, \hdots, K \},$ are convex and smooth. Consequently, they are also differentiable.}
%\vspace{2pt}
%\vspace{2pt}
%\noindent \textbf{Assumption \Diff. (Differentiability)} \textit{The functions $f_k(\cdot),$ $k \in \{1, \hdots, K \},$ are differentiable.}

Using \eqref{LagrangianReformulated}, and under Assumption $\Convex$, the steps of the PGFL algorithm without inter-cluster learning can be expressed as follows: %, where the Lagrangian $\Lmc_{\rho}$ is minimized with respect to $\wbf$, $\ebf$, and $\boldsymbol{\lambda}$,
% Note, we reintroduce privacy here. Those are basic private ADMM steps

\begin{align}
    \label{3StepLagrangian}
    \nabla F(\wbf^{(n+1)}) + {\bf A}^\mathsf{T} \boldsymbol{\lambda}^{(n)} + \rho {\bf A}^\mathsf{T} \Bigl( {\bf A} \wbf^{(n+1)} + {\bf B} \ebf^{(n)} \Bigr) = 0, \notag \\
    {\bf B}^\mathsf{T} \boldsymbol{\lambda}^{(n)} + \rho {\bf B}^\mathsf{T} \Bigl( {\bf A} \widetilde{\wbf}^{(n+1)} + {\bf B} \ebf^{(n+1)} \Bigr) = 0, \notag \\
    \boldsymbol{\lambda}^{(n+1)} - \boldsymbol{\lambda}^{(n)} + \rho \Bigl( {\bf A} \widetilde{\wbf}^{(n+1)} + {\bf B} \ebf^{(n+1)} \Bigr) = 0.
\end{align}
%The work in \cite{shi2014linear} proposes to reduce the above to two iterations by introducing the following matrices,
Similarly to \cite{shi2014linear}, we introduce the following to simplify \eqref{3StepLagrangian}:
%matrices to reduce the above equation to two steps.
\begin{equation*}
\begin{aligned}[c]
    {\bf H}_+ &= {\bf A}_1^\mathsf{T} + {\bf A}_2^\mathsf{T}, \notag \\
    {\bf L}_+ &= \frac{1}{2} {\bf H}_+ {\bf H}_+^\mathsf{T}, \notag \\
    \boldsymbol{\alpha} &= {\bf H}_-^\mathsf{T} \wbf, \notag \\
\end{aligned}
\qquad
\begin{aligned}[c]
    {\bf H}_- &= {\bf A}_1^\mathsf{T} - {\bf A}_2^\mathsf{T}, \notag \\
    {\bf L}_- &= \frac{1}{2} {\bf H}_- {\bf H}_-^\mathsf{T}, \notag \\
    {\bf M} &= \frac{1}{2} ({\bf L}_+ + {\bf L}_-). \notag \\
\end{aligned}
\end{equation*}
%We note that ${\bf L}_+$ and ${\bf L}_-$ correspond to the signless Laplacian and signed Laplacian matrices of the network seen as fully distributed, respectively. Hence, ${\bf L}_-$ is positive semi-definite with the nullspace given by $Null({\bf L}_-) = span\{{\bf 1}\}$.
Then, as derived in \cite[Section II.B]{shi2014linear}, \eqref{3StepLagrangian} becomes
\begin{align}
    \label{2StepLagrangian}
    \nabla F(\wbf^{(n+1)}) + \boldsymbol{\alpha}^{(n)} + 2 \rho {\bf M} \wbf^{(n+1)} - \rho {\bf L}_+ \widetilde{\wbf}^{(n)} &= 0, \notag \\
    \boldsymbol{\alpha}^{(n+1)} - \boldsymbol{\alpha}^{(n)} - \rho {\bf L}_- \widetilde{\wbf}^{(n+1)} &= 0.
\end{align}
%This formulation can be reduced to a single step using the work in \cite{li2017robust}. For this purpose, we introduce

As in \cite[Lemma 1]{li2017robust}, the equations in \eqref{2StepLagrangian} can be combined to obtain
%Combining the equations in \eqref{2StepLagrangian}, as in \cite[Lemma 1]{li2017robust}, we obtain
\begin{align}
    \label{li2017robusteq}
    \wbf^{(n+1)} = &\frac{{\bf M}^{-1}  \nabla F(\wbf^{(n+1)})}{2 \rho} + \frac{{\bf M}^{-1} {\bf L}_+ \widetilde{\wbf}^{(n)}}{2} \notag\\
    &- \frac{{\bf M}^{-1} {\bf L}_-}{2} \sum_{s=0}^{n} \widetilde{\wbf}^{(s)}.
\end{align}
Similarly to \cite{li2017robust}, by introducing the following:
%matrices to simplify the above equation:
\begin{equation*}
\begin{aligned}[c]
    {\bf Q} = \sqrt{{\bf L}_- / 2}, \notag \\
    {\bf q}^{(n)} = \dbinom{{\bf r}^{(n)}}{\widetilde{\wbf}^{(n)}}, \notag \\
\end{aligned}
\qquad
\begin{aligned}[c]
    {\bf r}^{(n)} = \sum_{s=0}^{n} {\bf Q} \widetilde{\wbf}^{(s)}, \notag \\
    {\bf G} = \begin{bmatrix}
        \rho {\bf I} & 0 \\
        0 & \rho {\bf L}_+ / 2
    \end{bmatrix},
\end{aligned}
\end{equation*}
%Note that by construction, $Null({\bf Q}) = span\{{\bf 1}\}$.
\eqref{li2017robusteq} can be reformulated using \cite[Lemma 2]{li2017robust} as
%and reformulate \eqref{li2017robusteq}, see \cite[Lemma 2]{li2017robust}, as
\begin{align}
    \label{1StepLagrangian}
    \frac{\nabla F(\wbf^{(n+1)})}{\rho} + 2 {\bf Q} {\bf r}^{(n+1)} + {\bf L}_+ \Bigl( \wbf^{(n+1)} - \widetilde{\wbf}^{(n)} \Bigr) = 2 {\bf M} \boldsymbol{\xi}^{(t+1)}.
\end{align}

\noindent \textbf{Theorem I.} \textit{Under Assumption $\Convex$, if $\tau^{(n)} = \tau = 0, \forall n$, the proposed PGFL algorithm converges to the optimal solution of \eqref{ERM_servers} in linear time for each cluster. }
\begin{proof}
Under Assumption $\Convex$, $F(\wbf)$ is convex and smooth by composition and, therefore, differentiable. Using \cite[Lemma 6]{Zhuhan} and \cite[Theorem V]{Zhuhan} with a convex and smooth function $F(\wbf)$ demonstrates that the proposed PGFL algorithm, without inter-cluster learning ($\tau = 0$ ), converges to the optimal solution of \eqref{ERM_servers} in linear time for any given cluster.
\end{proof}

% ZhuHan says it converges to an optimal solution of the mininzation problem

%%%%%%%%%%%%%%%%%%%%%%%%%%%%%%%%%%%%%%%%
\subsection{Impact of Inter-Cluster Learning}

In situations with limited data, as demonstrated in Section V, employing inter-cluster learning ($\tau \neq 0$) can enhance performance compared to $\tau = 0$. This section establishes an upper bound on the disparity between the resulting cluster-specific personalized models obtained in scenarios with and without inter-cluster learning. It is worth noting that this bound can be controlled by properly choosing the sequence $\tau(n)$.

%In situations with limited data, as demonstrated in Section V, employing inter-cluster learning ($\tau \neq 0$) can enhance performance compared to $\tau = 0$. Therefore, this section establishes an upper bound on the disparity between the resulting cluster-specific personalized models obtained in scenarios with and without inter-cluster learning. It is worth noting that this bound can be controlled by properly choosing the sequence $\tau(n)$.
%In cases were data is limited, {\color{red}the optimal solution of \eqref{ERM_servers} with $\tau = 0$ may be inferior to that obtained with $\tau \neq 0$.} In this section, we bound the difference between the resulting cluster-specific personalized models with and without inter-cluster learning. Notably, the obtained bound can be adjusted by modifying the sequence $\tau^{(n)}$, and depends on the cluster similarity.
%This ensures that the proposed PGFL algorithm with inter-cluster learning converges, for each cluster, to a neighborhood of the optimal cluster-specific solution. 

To do so, it is necessary to reformulate the client primal update using Assumption $\Convex$. The primal update for client $k \in \Cmc_{s, (q)}$ is expressed as follows:
\begin{align}
    \wbf_k^{(n+1)} = &\text{arg} \min_{\wbf} f_k(\wbf) - \left< \varphibf_k^{(n)}, \wbf - \wbf_{q,s}^{(n)} \right> \notag \\
    &+ \frac{\rho}{2} ||\wbf - \wbf_{q,s}^{(n)}||^2,
\end{align}
which, under Assumption $\Convex$, is equivalent to
\begin{align}
    \nabla f_k(\wbf_k^{(n+1)}) - \varphibf_k^{(n)} + \rho \Bigl( \wbf_k^{(n+1)} - \wbf_{q,s}^{(n)} \Bigr) = 0.
\end{align}
Further reformulation leads to the following:
\begin{align}
    \label{SimpleClientUpdate}
    \wbf_k^{(n+1)} = \wbf_{q,s}^{(n)} + \frac{1}{\rho} \varphibf_k^{(n)} - \frac{1}{\rho} \nabla f_k(\wbf_k^{(n+1)}).
\end{align}

%Recall the definition of the server aggregation \eqref{ServerAggregation}:
%\begin{align}
%    \widehat{\wbf}_{q,s}^{(n)} &= \frac{1}{|\Nmc_s|} \sum_{\sprime \in \Nmc_s} \widetilde{\wbf}_{q, \sprime}^{(n)}, \notag \\
%    &= \frac{1}{|\Nmc_s|} \sum_{\sprime \in \Nmc_s} \frac{1}{|\Cmc_{\sprime, (q)}|} \sum_{k \in \Cmc_{\sprime, (q)}} \Bigl( \wbf_k^{(n)} - \frac{1}{\rho} \varphibf_k^{(n-1)} \Bigr),
%\end{align}
%which, using \eqref{SimpleClientUpdate}, can be reformulated as
By replacing $\wbf_k^{(n+1)}$ with \eqref{SimpleClientUpdate} in \eqref{ServerAggregation}, we obtain
\begin{align}
    \label{NewExpressionServerModel}
    \widehat{\wbf}_{q,s}^{(n)} &= \frac{1}{|\Nmc_s|} \sum_{\sprime \in \Nmc_s} \frac{1}{|\Cmc_{\sprime, (q)}|} \sum_{k \in \Cmc_{\sprime, (q)}} \Bigl( \wbf_{q, \sprime}^{(n-1)} - \frac{1}{\rho} \nabla f_k(\wbf_k^{(n)}) \Bigr).
\end{align}

Next, we investigate the effect of inter-cluster learning by comparing the performance of models obtained using the PGFL algorithm with and without inter-cluster learning. We shall prove that the difference between the resulting models is bounded and depends on both the inter-cluster learning parameter and the similarity of models between clusters.
% For this purpose, we refer to these models as $\bar{\wbf}_k$ and $\bar{\wbf}_{q,s}$ for the client and server, respectively.
%To evaluate the impact of inter-cluster learning, we compare the models obtained using the proposed PGFL algorithm with and without inter-cluster learning. In the following, we denote such models $\bar{\wbf}_k$ and $\bar{\wbf}_{q,s}$ for client and server, respectively. We will prove that the difference between the resulting models is bounded and that the bound is a function of the inter-cluster learning parameter and the model similarity between clusters. To this aim, it is necessary to quantify the latter with the following definition.

\vspace{2pt}

\noindent \textbf{Theorem II.} \textit{Given a sufficiently large penalty parameter $\rho$, for all iterations, server $s \in \Smc$ and cluster $q \in \Qmc$, the impact of inter-cluster learning after $n$ iterations is bounded by }
\begin{align}
    \label{Recursion}
    \Ebb \Bigl[ || \bar{\wbf}_{q,s}^{(n)} - \wbf_{q,s}^{(n)} ||_2^2 \Bigr] \leqslant \sum_{i = 1}^n \Bigl( \prod_{j = i+1}^n \bigl( 1 - \tau^{(j)} \bigr) \Bigr) \tau^{(i)} \eta,
\end{align}
%\textcolor{red}{the expectation is taken with respect to the privacy-related and observation noises}
\textit{where the expectation is taken with respect to the privacy-related noise added in \eqref{PrivacyNoiseAddition} and the data observation noise, $\bar{\wbf}_{q,s}^{(n)}$ denotes the model obtained by the algorithm without inter-cluster learning, and $\eta$ is the maximum cluster model distance, defined as:
\begin{align}
    \eta = \max_{q,r \in \Qmc} \norm{\wbf_q^* - \wbf_r^*}_2^2,
\end{align}
with the models $\wbf_q^*, q \in \Qmc$ being the cluster-specific solutions of \eqref{ERM_servers} with $\tau = 0$.}
\begin{proof}
We prove this theorem by induction. With initial values $\wbf_{q,s}^{(0)} = {\bf 0}$ and $\bar{\wbf}_{q,s}^{(0)} = {\bf 0}$, one can write. 
\begin{align}
    \wbf_{q,s}^{(1)} &= \Bigl( 1 - \tau^{(1)} \Bigr) \widehat{\wbf}_{q,s}^{(1)} + \frac{\tau^{(1)}}{Q-1} \sum_{r \in \Qmc \backslash q} \widehat{\wbf}_{r,s}^{(1)}, \notag \\
    \bar{\wbf}_{q,s}^{(1)} &= \frac{1}{|\Nmc_s|} \sum_{\sprime \in \Nmc_s} \frac{1}{|\Cmc_{\sprime, (q)}|} \sum_{k \in \Cmc_{\sprime, (q)}} \Bigl( \bar{\wbf}_{q, \sprime}^{(0)} - \frac{1}{\rho} \nabla f_k(\bar{\wbf}_{k}^{(1)}) \Bigr),
\end{align}
where, given that $\bar{\wbf}_{q, \sprime}^{(0)} = \wbf_{q, \sprime}^{(0)}$ and $\bar{\wbf}_{k}^{(0)} = \wbf_{k}^{(0)}$, and using \eqref{NewExpressionServerModel}, we have $\widehat{\wbf}_{q,s}^{(1)} = \bar{\wbf}_{q,s}^{(1)}$. Hence,
\begin{align}
    \bar{\wbf}_{q,s}^{(1)} - \wbf_{q,s}^{(1)} = \frac{\tau^{(1)}}{Q-1} \sum_{r \in \Qmc \backslash q} \Bigl( \bar{\wbf}_{q,s}^{(1)} - \widehat{\wbf}_{r,s}^{(1)} \Bigr).
\end{align}
Taking the expectation with respect to the privacy-related and observation noises, we can express this difference as a function of the inter-cluster learning parameter and the maximum cluster model distance.
\begin{align}
    \Ebb [ || \bar{\wbf}_{q,s}^{(1)} - \wbf_{q,s}^{(1)} ||_2^2 ] \leqslant \tau^{(1)} \eta.
\end{align}

Further, we assume that \eqref{Recursion} is satisfied for all iterations up to iteration $n-1$. For iteration $n$, we have
\begin{align}
    \wbf_{q,s}^{(n)} &= \Bigl( 1 - \tau^{(n)} \Bigr) \widehat{\wbf}_{q,s}^{(n)} + \frac{\tau^{(n)}}{Q-1} \sum_{r \in \Qmc \backslash q} \widehat{\wbf}_{r,s}^{(n)}, \notag \\
    \bar{\wbf}_{q,s}^{(n)} &= \frac{1}{|\Nmc_s|} \sum_{\sprime \in \Nmc_s} \frac{1}{|\Cmc_{\sprime, (q)}|} \sum_{k \in \Cmc_{\sprime, (q)}} \Bigl( \bar{\wbf}_{q, \sprime}^{(n-1)} - \frac{1}{\rho} \nabla f_k(\bar{\wbf}_k^{(n)}) \Bigr),
\end{align}
where $\widehat{\wbf}_{q,s}^{(n)} \neq \bar{\wbf}_{q,s}^{(n)}$ since 
\begin{align}
    \widehat{\wbf}_{q,s}^{(n)} &= \frac{1}{|\Nmc_s|} \sum_{\sprime \in \Nmc_s} \frac{1}{|\Cmc_{\sprime, (q)}|} \sum_{k \in \Cmc_{\sprime, (q)}} \Bigl( \wbf_{q, \sprime}^{(n-1)} - \frac{1}{\rho} \nabla f_k(\wbf_k^{(n)}) \Bigr).
\end{align}
The difference is given by
\begin{align}
    \label{Difference}
    \bar{\wbf}_{q,s}^{(n)} - &\wbf_{q,s}^{(n)} = \Bigl( 1 - \tau^{(n)} \Bigr) \Bigl( \bar{\wbf}_{q,s}^{(n)} - \widehat{\wbf}_{q,s}^{(n)} \Bigr) \notag \\
    &+ \frac{\tau^{(n)}}{Q-1} \sum_{r \in \Qmc \backslash q} \Bigl( \bar{\wbf}_{q,s}^{(n)} - \widehat{\wbf}_{r,s}^{(n)} \Bigr),
\end{align}
with
\begin{align}
    \label{useAssumption2}
    \bar{\wbf}_{q,s}^{(n)} - \widehat{\wbf}_{q,s}^{(n)} = &\frac{1}{|\Nmc_s|} \sum_{\sprime \in \Nmc_s} \frac{1}{|\Cmc_{\sprime, (q)}|} \sum_{k \in \Cmc_{\sprime, (q)}} \Bigl( \bar{\wbf}_{q, \sprime}^{(n-1)} \notag \\
    & - \wbf_{q, \sprime}^{(n-1)} - \frac{1}{\rho} \nabla f_k(\bar{\wbf}_k^{(n)}) + \frac{1}{\rho} \nabla f_k(\wbf_k^{(n)}) \Bigr).
\end{align}
We note that the expectation of $|| \bar{\wbf}_{q, \sprime}^{(n-1)} - \wbf_{q, \sprime}^{(n-1)} ||_2^2$ with respect to the privacy-related and observation noises is identical for all servers. Therefore, since \eqref{Recursion} is satisfied for iteration $n-1$ for all servers, given a sufficiently large penalty parameter $\rho$, and taking the expectation with respect to the privacy-related and observation noises, we have
% can discuss rho further here, as only the difference in gradients is ignored
\begin{align}
    \label{DifferenceAggregate}
    \Ebb || \bar{\wbf}_{q,s}^{(n)} - \widehat{\wbf}_{q,s}^{(n)} ||_2^2 \leqslant \Ebb || \bar{\wbf}_{q,s}^{(n-1)} - \wbf_{q,s}^{(n-1)} ||_2^2.
\end{align}
% Previous eq, can instead put on the right \eqref{Recursion}
Combining \eqref{Difference} and \eqref{DifferenceAggregate}, we will have
\begin{align}
    \Ebb || \bar{\wbf}_{q,s}^{(n)} - \wbf_{q,s}^{(n)} ||_2^2 &\leqslant (1 - \tau^{(n)}) \Ebb || \bar{\wbf}_{q,s}^{(n-1)} - \wbf_{q,s}^{(n-1)} ||_2^2 \notag \\
    & + \frac{\tau^{(n)}}{Q-1} \sum_{r \in \Qmc \backslash q} \Ebb || \bar{\wbf}_{q,s}^{(n)} - \widehat{\wbf}_{r,s}^{(n)} ||_2^2,
\end{align}
which, using the maximum cluster model distance, yields
\begin{align}
    \Ebb || \bar{\wbf}_{q,s}^{(n)} - \wbf_{q,s}^{(n)} ||_2^2 &\leqslant \Bigl( 1 - \tau^{(n)} \Bigr) \Ebb || \bar{\wbf}_{q,s}^{(n-1)} - \wbf_{q,s}^{(n-1)} ||_2^2 \notag \\
    & + \tau^{(n)} \eta.
\end{align}
Given \eqref{Recursion} for iteration $n-1$, we have
\begin{align}
    \Ebb || \bar{\wbf}_{q,s}^{(n)} - \wbf_{q,s}^{(n)} ||_2^2 &\leqslant \Bigl( 1 - \tau^{(n)} \Bigr) \sum_{i = 1}^{n-1} \Bigl( \prod_{j = i+1}^{n-1} (1 - \tau^{(j)}) \Bigr) \tau^{(i)} \eta \notag \\
    & + \tau^{(n)} \eta, \notag \\
    &\leqslant \sum_{i = 1}^n \Bigl( \prod_{j = i+1}^n (1 - \tau^{(j)}) \Bigr) \tau^{(i)} \eta.
\end{align}
\noindent That is, \eqref{Recursion} is satisfied for iteration $n$.

By the principle of induction, \eqref{Recursion} is satisfied for all iterations, server $s \in \Smc$ and cluster $q \in \Qmc$.

%By mathematical induction, since the inequality \eqref{Recursion} is satisfied at iteration $1$, and that it being satisfied at iteration $k$ implies that it is satisfied at iteration $k+1$, it is satisfied at all iterations. 
\end{proof}

\noindent \textbf{Corollary.} \textit{If $\tau^{(i)} = 0, \forall i < n$ and $\tau^{(n)} \neq 0$, the impact of a single iteration of inter-cluster learning is bounded by }
\begin{align}
    \Ebb || \bar{\wbf}_{q,s}^{(n)} - \wbf_{q,s}^{(n)} ||_2^2 \leqslant \tau^{(n)} \eta,
\end{align}
\textit{where $\bar{\wbf}_{q,s}^{(n)}$ denotes a model obtained without inter-cluster learning, $\eta$ is as defined in Theorem II, and the expectation is taken with respect to the privacy-related and observation noises.}

Theorem II bounds the difference in the resulting models with and without inter-cluster learning. %Combined with Theorem I, this ensures that the models obtained by the algorithms are within the neighborhood of the optimal solution of \eqref{ERM_servers} with $\tau = 0$. When the participating clients have access to sufficient data, this solution is satisfactory, and the algorithm converges to a neighborhood of it. Furthermore, the size of this neighborhood can be adjusted by selecting the sequence $\tau^{(n)}$. When data is scarce, however, the solution of \eqref{ERM_servers} with $\tau = 0$ may be unsatisfactory, and inter-cluster learning enables the proposed algorithm to achieve better accuracy, as we will observe in Section V.
Combining Theorems I and II, the resulting models obtained by the algorithms are guaranteed to reside within a neighborhood of the optimal solution of \eqref{ERM_servers} with $\tau = 0$. The size of this neighborhood can be adjusted by selecting the sequence $\tau^{(n)}$. When ample data is available, the algorithm converges to a satisfactory solution within this neighborhood. However, in cases of limited data, the solution of \eqref{ERM_servers} with $\tau = 0$ may be inadequate. In such situations, inter-cluster learning becomes crucial, allowing the proposed algorithm to achieve higher accuracy, as demonstrated in Section V. By exploiting inter-cluster learning, the algorithm effectively overcomes the limitations imposed by scarce data, leading to improved performance. 

%We have shown that the resulting model difference with the use of inter-cluster learning is bounded. This implies that when each cluster has enough data to ensure convergence to a satisfactory model, the proposed PGFL algorithm converges to a neighborhood of this solution. Furthermore, careful selection of the sequence $\tau^{(n)}$ can reduce the size of this neighborhood. When data is scarce, inter-cluster learning can increase accuracy, as we will observe in Section V.

%This provides a bound for the impact of inter-cluster learning on the accuracy of the proposed PGFL.
% Cut following Stefan's comment: Note that the impact of inter-cluster learning is often positive if controlled properly, especially when data is scarce, as we will see in Section V on the MNIST and synthetic datasets.

%%%%%%%%%%%%%%%%%%%%%%%%%%%%%%%%%%%%%%%%%%%%%%%%%%%%%%%%%%%%%%%%%%%%%%%%%%%%%%%%
\section{Privacy Analysis}
This section focuses on quantifying the local privacy protection provided by the proposed PGFL algorithm. To achieve this, we begin by calculating the $l_2$-norm sensitivity, which quantifies the variation in output resulting from a change in an individual data sample. Once we  have established the $l_2$-norm sensitivity, we proceed to adjust the noise variance added to the primal variables, ensuring satisfactory protection.
%In this section, the local privacy protection of the proposed PGFL algorithm is quantified. For this purpose, we first compute the $l_2$-norm sensitivity, which measures the difference in output when an individual data sample is changed. Once the $l_2$-norm sensitivity is established, it will be possible to calibrate the variance of the noise added to the primal variables to guarantee satisfactory protection. 
\\

\noindent \textbf{Definition 1.} \textit{The $l_2$-norm sensitivity is defined by}
\begin{equation}
\Delta_{k,2}^{(n)} = \max_{\mathcal{D}_k,\mathcal{D}_l} \norm{ \wbf_{k,\mathcal{D}_k}^{(n)} - \wbf_{k,\mathcal{D}_l}^{(n)}}
\end{equation}
\textit{where $\wbf_{k,\mathcal{D}_k}^{(n)}$ and $\wbf_{k,\mathcal{D}_l}^{(n)}$ denote the local primal variables obtained from two neighboring data sets $\mathcal{D}_k$ and $\mathcal{D}_l$, which differ in only one data sample.}

\vspace{2pt}
\noindent \textbf{Assumption \Grad.} \textit{The functions $\ell_k(\cdot),$ $k \in \Cmc,$ have bounded gradients. That is, for $k \in \Cmc$ there exists a constant $C_k$ such that $||\nabla \ell_k(\cdot)|| \leqslant C_k$.}
\vspace{2pt}

\noindent \textbf{Lemma 1.} \textit{Under Assumption $\Grad$, the $l_2$-norm sensitivity for a client $k$ is given by
\begin{equation}
    \Delta_{k,2}^{(n)} = \max_{\mathcal{D}_k,\mathcal{D}_l} || \wbf_{k,\mathcal{D}_k}^{(n)} - \wbf_{k,\mathcal{D}_l}^{(n)} || = \frac{2 C_k}{\rho D_k}.
\end{equation}}

\begin{proof}
%See Appendix A.
We consider two neighboring data sets for a client $k$, $\mathcal{D}_k$ and $\mathcal{D}_l$, both of cardinality $D_k$. For simplicity, we assume that they differ on the last data sample. We denote $\wbf_{k,\mathcal{D}_k}^{(n)}$ the model obtained using the initial data set, and $\wbf_{k,\mathcal{D}_l}^{(n)}$ the model obtained using the alternative data set. Those are obtained, according to \eqref{ClientPrimalUpdate}, by:

\begin{align}
    \wbf_{k,\mathcal{D}_k}^{(n)} &= \text{arg} \min_{\wbf} \frac{1}{D_k} \sum_{i=1}^{D_k} \ell_k({\bf x}_{k,i}, y_{k,i}; \wbf) + \frac{\lambda}{|\Cmc_s|} R(\wbf) \notag \\
    &- \left< \varphibf_k^{(n-1)}, \wbf - \wbf_{q,s}^{(n-1)} \right> + \frac{\rho}{2} ||\wbf - \wbf_{q,s}^{(n-1)}||^2, \notag \\
    \wbf_{k,\mathcal{D}_l}^{(n)} = &\text{arg} \min_{\wbf} \frac{\lambda}{|\Cmc_s|} R(\wbf) \notag \\
    &+ \frac{1}{D_k} \Bigl( \sum_{i=1}^{D_k-1} \ell_k({\bf x}_{k,i}, y_{k,i}; \wbf) + \ell_k({\bf x}_{k,D_k}', y_{k,D_k}'; \wbf) \Bigr) \notag \\
    &- \left< \varphibf_k^{(n-1)}, \wbf - \wbf_{q,s}^{(n-1)} \right> + \frac{\rho}{2} ||\wbf - \wbf_{q,s}^{(n-1)}||^2. \notag
\end{align}

Using \eqref{SimpleClientUpdate}, that we recall:
\begin{align}
    \wbf_k^{(n)} = \wbf_{q,s}^{(n-1)} + \frac{1}{\rho} \varphibf_k^{(n-1)} - \frac{1}{\rho} \nabla f_k(\wbf_k^{(n)}),
\end{align}
we can derive:
\begin{align}
    &|| \wbf_{k,\mathcal{D}_k}^{(n)} - \wbf_{k,\mathcal{D}_l}^{(n)} || = \\
    &\norm{ \frac{1}{\rho D_k} ( \nabla \ell_k({\bf x}_{k,D_k}, y_{k,D_k}; \wbf_k) - \nabla \ell_k({\bf x}_{k,D_k}', y_{k,D_k}'; \wbf_k) ) }, \notag
\end{align}
which, under Assumption $\Grad$, provides a value for the $l_2$-norm sensitivity:
\begin{align}
    \max_{\mathcal{D}_k,\mathcal{D}_l} || \wbf_{k,\mathcal{D}_k}^{(n)} - \wbf_{k,\mathcal{D}_l}^{(n)} || = \frac{2 C_k}{\rho D_k}.
\end{align}
\end{proof}

With the $l_2$-norm sensitivity, we can establish the relation between the noise variance added in \eqref{PrivacyNoiseAddition} and the privacy parameter $\phi_k^{(n)}$ as well as prove the privacy guarantee of the algorithm in terms of zCDP. \\

\noindent \textbf{Theorem III.} \textit{Under Assumption $\Grad$, PGFL satisfies dynamic $\phi_k^{(n)}$-zCDP with the relation between the privacy parameter and the perturbation noise variance given by}
\begin{align}
    \delta_k^{2(n)} = \frac{\Delta_{k,2}^{(n) 2}}{2 \phi_k^{(n)}}.
\end{align}

\begin{proof}
%See Appendix B.
For any client $k$ and iteration $n$, the perturbed primal update is obtained with \eqref{PrivacyNoiseAddition}. That is, it is equivalent to $\widetilde{\wbf}_k^{(n)} \sim \Nmc(\wbf_k^{(n)}, \delta_k^{2(n)} \Ibf)$. Hence, for two neighboring data sets $\mathcal{D}_k$ and $\mathcal{D}_l$, we have $\widetilde{\wbf}_{k,\mathcal{D}_k}^{(n)} \sim \Nmc(\wbf_{k,\mathcal{D}_k}^{(n)}, \delta_k^{2(n)} \Ibf)$ and $\widetilde{\wbf}_{k,\mathcal{D}_l}^{(n)} \sim \Nmc(\wbf_{k,\mathcal{D}_l}^{(n)}, \delta_k^{2(n)} \Ibf)$

Using \cite[Lemma 17]{Bun}, which states $D_{\alpha}(N(\mu,v \boldsymbol{I}_d) || N(\nu,v \boldsymbol{I}_d)) = \frac{\alpha ||\mu - \nu ||_2^2}{2 v}$, $\forall \alpha \in [1,\infty)$; we obtain, $\forall \alpha \in [1,\infty)$, the following Kullback-Leibler-divergence:
\begin{align}
    D_{\alpha}(\widetilde{\wbf}_{k,\mathcal{D}_k}^{(n)} || \widetilde{\wbf}_{k,\mathcal{D}_l}^{(n)}) = \frac{\alpha || \wbf_{k,\mathcal{D}_k}^{(n)} - \wbf_{k,\mathcal{D}_l}^{(n)} ||_2^2}{2 \delta_k^{2(n)}}.
\end{align}

Using Lemma 1, we can bound the KL-divergence by
\begin{align}
    \label{eqprivproof}
    D_{\alpha}(\widetilde{\wbf}_{k,\mathcal{D}_k}^{(n)} || \widetilde{\wbf}_{k,\mathcal{D}_l}^{(n)}) \leqslant \frac{\alpha \Delta_{k,2}^{(n) 2}}{2 \delta_k^{2(n)}}.
\end{align}

Further, we consider the privacy loss of $\widetilde{\wbf}_{k}^{(n)}$ at output $\lambda$:
\begin{align}
   \boldsymbol{z}_k^{(n)}(\widetilde{\wbf}_{k,\mathcal{D}_k}^{(n)} || \widetilde{\wbf}_{k,\mathcal{D}_l}^{(n)}) = \log \frac{P(\widetilde{\wbf}_{k,\mathcal{D}_k}^{(n)} = \lambda)}{P(\widetilde{\wbf}_{k,\mathcal{D}_l}^{(n)} = \lambda)}.
\end{align}

%Given $D_{\alpha}(\cdot) \leqslant \epsilon + \rho \alpha \Longleftrightarrow E(e^{(\alpha-1)Z(\cdot)}) \leqslant e^{(\alpha - 1)(\epsilon + \rho \alpha)}$, we have
Using the definition of the KL-divergence with \eqref{eqprivproof}, we obtain
\begin{align}
    E(e^{(\alpha - 1)\boldsymbol{z}_k^{(n)}(\lambda)}) &\leqslant e^{(\alpha - 1) D_{\alpha}(\widetilde{\wbf}_{k,\mathcal{D}_k}^{(n)} || \widetilde{\wbf}_{k,\mathcal{D}_l}^{(n)})} \notag \\
    &\leqslant e^{(\alpha - 1) \frac{\alpha \Delta_{k,2}^{(n) 2}}{2 \delta_k^{2(n)}}}.
\end{align}

Thus, the PGFL algorithm satisfies the dynamic $\phi_{k}^{(n)}$-zCDP with $\phi_{k}^{(n)} = \frac{\Delta_{k,2}^{(n)}}{2 \delta_k^{2(n)}}$.
\end{proof}

Theorem III gives the relationship between the noise perturbation variance and the privacy protection at a given iteration. Given that the proposed algorithm is an iterative process and several estimates are exchanged, one needs to consider the total privacy loss throughout the learning process. The total privacy loss after $n$ iterations can be computed using \cite[Theorem 3]{Zhuhan} and is given in terms of $(\epsilon, \delta)$-DP for any $\decreaserate \in (0,1)$ and $\delta \in (0,1)$ by
\begin{align}
    \epsilon = \max_{k \in \Cmc} \Biggl( \frac{\phi_{k}^{(1)} (1 - \decreaserate^n)}{\decreaserate^{n-1} - \decreaserate^n} + 2 \sqrt{ \frac{\phi_{k}^{(1)} (1 - \decreaserate^n) \ln 1/\delta}{\decreaserate^{n-1} - \decreaserate^n} } \Biggr).
\end{align}
%Given a total privacy leakage bound and a final iteration, one can calculate the variance of the noise perturbation using the above theorem.
%The above theorem enables the calibration of the variance of the noise perturbation given a total privacy leakage bound and a final iteration.

%%%%%%%%%%%%%%%%%%%%%%%%%%%%%%%%%%%%%%%%%%%%%%%%%%%%%%%%%%%%%%%%%%%%%%%%%%%%%%%%
\section{Numerical Simulations}

This section illustrates the performance of the proposed PGFL algorithm for solving regression and classification tasks.
%synthetic data and on the MNIST dataset \cite{MNIST}.

%%%%%%%%%%%%%%%%%%%%%%%%%%%%%%%%%%%%%%%%
%\subsection{Simulation Setup for Regression}
\subsection{Experiments for Regression}

We consider a graph federated network consisting of $|\Smc| = 10$ servers, each having access to $|\Cmc_s| = 15$ clients, for a total of $|\Cmc| = 150$ clients. The set of servers and their communication channels form a random connected graph where the average node degree is three. Each client has access to a random number of noisy data samples between $D_k = 2$ and $D_k = 9$, each composed of a vector ${\bf x}_{k,i}$ of dimension $d = 60$ and a response scalar $y_{k,i}$. Doing so, each cluster is globally observable but not locally at any given client or set $\Cmc_s, s \in \Smc$. The servers implement random scheduling of clients to reduce the communication load \cite{FedAVG}. In particular, at every global iteration, each server randomly selected a subset of three clients to participate in the learning process.

The clients of the network are randomly split between $Q = 3$ clusters. Clients of a given cluster solve the ridge regression problem with data generated from an original model $\wbf_q^*$, obtained with $\wbf_q^* = \wbf_0^* + \gamma \wbf_0^*$ with $\gamma \sim \Umc(-0.15,0.15)$, where $\wbf_0^*$ is a base model. In doing so, the learning tasks of the different clusters share the same objective functions but have different, related data distributions. The loss and regularizer functions are given by
\begin{align}
    \ell_k({\bf X}_k, {\bf y}_k; \wbf_k) &= ||{\bf y}_k - {\bf X}_k \wbf_k||^2, \notag \\
    R(\wbf_k) &= ||\wbf_k||^2.
\end{align}

Performance is evaluated by computing the normalized mean squared deviation (NMSD) of the local models with respect to the corresponding cluster-specific original model used to generate the data, $\wbf_q^*$ for $k \in \Cmc_{(q)}$. It is given by:
\begin{align}
    \label{MSE}
    \gamma^{(n)} =\frac{1}{|\Cmc|} \sum_{q=1}^{|\Qmc|} \sum_{k \in \Cmc_{(q)}} \frac{ \norm{ \wbf_{k}^{(n)} - \wbf_q^*}_2^2}{ \norm{\wbf_q^*}_2^2},
\end{align}
where the result is averaged over several Monte Carlo iterations. To ensure a fair comparison, the algorithms are set to observe the same initial convergence rate whenever possible. For most experiments, we display the learning curve, that is, the NMSD against the iteration index.

%%%%%%%%%%%%%%%%%%%%%%%%%%%%%%%%%%%%%%%%
%\subsection{Experiments for Regression}

%\begin{figure*}[t!]
%    \centering
%    \subfigure[]{\includegraphics[width=0.32\textwidth]{Figures/Journal/PGFL_Main_Basic.eps}}
%    \subfigure[]{\includegraphics[width=0.32\textwidth]{Figures/Journal/PGFL_Main_Sched.eps}}
%    \subfigure[]{\includegraphics[width=0.32\textwidth]{Figures/Journal/PGFL_Main_SchedPriv.eps}}
%    \caption{Performance of the PGFL algorithm on a regression task, with FedAvg simulated for comparison purposes. Learning curves (a) without client scheduling nor privacy, (b) with client scheduling, (c) with client scheduling and privacy. }
%    \label{PGFL_main}
%\end{figure*}

% Basic
We first consider an ideal setting wherein all algorithms are evaluated without privacy considerations ($\xibf^{(n)} = {\bf 0}, \; \forall n$)) and client scheduling. Then, for comparison purposes, we adapted the conventional federated averaging (FedAvg) algorithm \cite{FedAVG}, learning a single global model, to the graph federated architecture. In this scenario, the inter-cluster parameter $\tau^{(n)}$ of the PGFL algorithm was kept fixed throughout the learning, specifically, $\tau^{(n)} = 0$ and $\tau^{(n)} = 0.4$, and did not employ inter-server communication when $\Emc = \emptyset$. 
Figure \ref{PGFL_main_a} shows the learning curves for the PGFL and FedAvg implementations above. The results illustrate the superiority of the proposed PGFL algorithm over FedAvg, as cluster-specific learning tasks benefit significantly from personalized models tailored to each cluster. We also see that incorporating inter-cluster learning results in improved convergence speed and steady-state accuracy. Furthermore, the performance of the PGFL algorithm is notably poor in the absence of inter-server communication, emphasizing the importance of using the graph federated architecture. Leveraging the model similarities improves learning speed and accuracy by compensating for data scarcity. In addition, isolated servers whose clients lack sufficient data to achieve satisfactory accuracy independently reinforce the necessity of the graph federated architecture.

% Client scheduling
Next, we modify the setting to incorporate client scheduling and evaluate the aforementioned algorithms with reduced communication load.
Figure \ref{PGFL_main_b} shows the learning curves for the PGFL and FedAvg with client scheduling. We observe that the PGFL algorithm exhibits slower convergence and higher steady-state NMSD when utilizing client scheduling. And we note that FedAvg performs better with client scheduling. The performance degradation for the PGFL algorithm is due to the lower client participation resulting in a smaller quantity of data being utilized. The better performance of FedAvg in this setting is due to the imbalance of cluster representation in the universal model, which benefits the participating clients on average.
%The slight increase in performance for the FedAvg algorithm is caused by the imbalance in cluster representation at each iteration, which benefits the participating clients on average.

\begin{figure}
    \centering
    \includegraphics[width=0.35\textwidth]{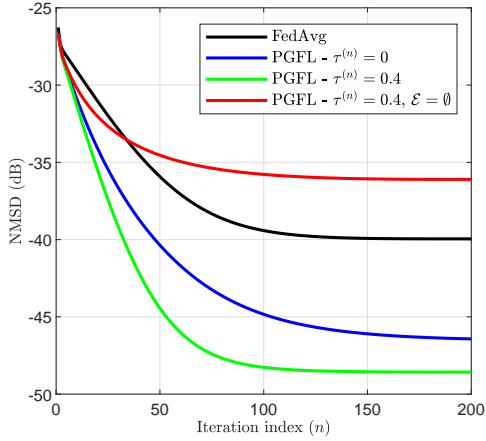}
    \caption{ Learning curves of the PGFL algorithm with a fixed inter-cluster learning parameter and the FedAvg algorithm, without client scheduling or privacy.} %Learning curves for ridge regression without client scheduling or privacy.}
    \label{PGFL_main_a}
\end{figure}

% Privacy
Finally, we evaluate the aforementioned algorithms in a setting with client scheduling and privacy protection. All of the algorithms utilize zCDP with the noise perturbation presented in \eqref{PrivacyNoiseAddition}.
Figure \ref{PGFL_main_c} shows the learning curves for the PGFL and FedAvg with client scheduling and privacy. We observe that the noise perturbation associated with differential privacy significantly reduces the convergence speed of all the simulated algorithms. However, we note that the NMSD after 300 iterations is nearly identical to the one in Fig. \ref{PGFL_main_b}. This behavior is explained by the use of zCDP, in which the variance of the noise perturbation starts high and decreases linearly throughout the learning process.

%\begin{figure*}[t!]
%    \centering
%    \subfigure[]{\includegraphics[width=0.32\textwidth]{Figures/Journal/PGFL_Taus.eps}}
%    \subfigure[]{\includegraphics[width=0.32\textwidth]{Figures/Journal/PGFL_TauDecrease.eps}}
%    \subfigure[]{\includegraphics[width=0.32\textwidth]{Figures/Journal/PGFL_TradeOff.eps}}
%    \caption{Performance of the PGFL algorithm on a regression task. (a) NMSD after $200$ iteration as a function of $\tau$, (b) learning curves with null, fixed, and decreasing $\tau$ in a setting with dissimilar tasks, and (c) privacy-accuracy trade-off. }
%    \label{PGFL_additional}
%\end{figure*}

\begin{figure}
    \centering
    \includegraphics[width=0.35\textwidth]{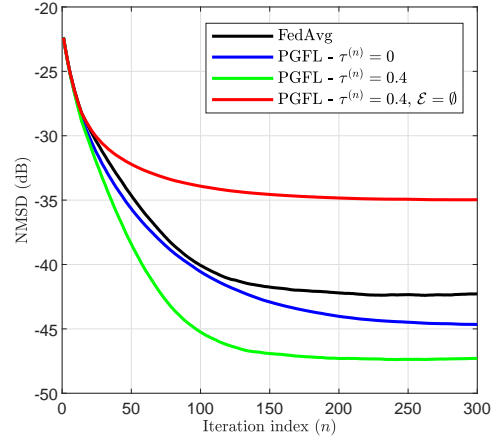}
    \caption{Learning curves of the PGFL algorithm with a fixed inter-cluster learning parameter and the FedAvg algorithm, considering client scheduling and without privacy.}
    \label{PGFL_main_b}
\end{figure}
\begin{figure}
    \centering
    \includegraphics[width=0.35\textwidth]{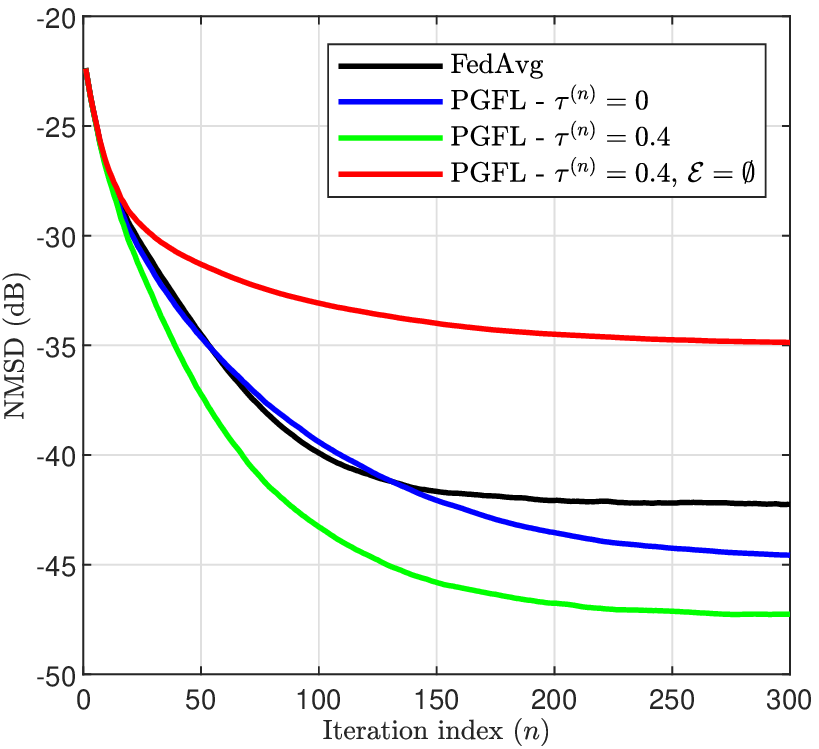}
    \caption{Learning curves of the PGFL algorithm with a fixed inter-cluster learning parameter and the FedAvg algorithm, considering client scheduling and privacy.}
    \label{PGFL_main_c}
\end{figure}
\begin{figure}
    \centering
    \includegraphics[width=0.35\textwidth]{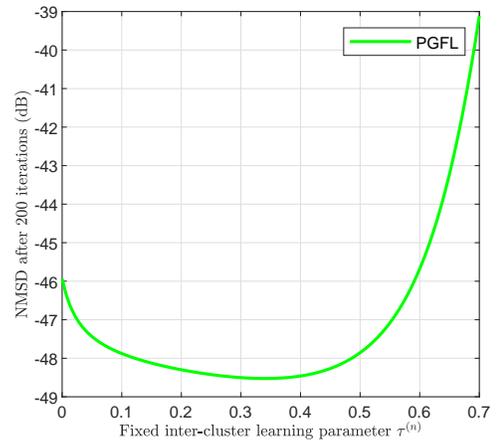}
    \caption{NMSD after $200$ iterations vs. fixed inter-cluster learning parameter $\tau^{(n)}$ values for the PGFL algorithm with client scheduling and privacy .}
    \label{PGFL_additional_a}
    \end{figure}

% Tau choice
Further, we illustrate the importance of carefully choosing the value of the inter-cluster learning parameter. In Fig. \ref{PGFL_additional_a}, we simulated the proposed PGFL algorithm for various fixed $\tau^{(n)}$ values and displayed the NMSD after $200$ iterations. For instance, the NMSD for $\tau^{(n)} = 0.4$ corresponds to the result obtained in Fig. \ref{PGFL_main_c}. This figure confirms that inter-cluster learning has the potential to increase learning performance by alleviating data scarcity, as the PGFL algorithm achieves lower NMSD with $\tau^{(n)} \in (0.1, 0.5)$ than with $\tau^{(n)} = 0$. It also shows that the inter-cluster learning parameter must be carefully selected, as a value too large for the setting leads to performance degradation.

% Decreasing taus
We then illustrate an alternative use of inter-cluster learning. For this experiment, the difference between the data distribution of the different clusters has been increased. Precisely, the datasets were simulated with the models obtained by $\wbf_q = \wbf_0 + \gamma \wbf_0$ with $\gamma \sim \Umc(-0.5,0.5)$. The learning curves are presented in Fig. \ref{PGFL_additional_b}. We observed that, because of the higher cluster dissimilarity, inter-cluster learning degrades steady-state NMSD; this is observed in the learning curves for PGFL with $\tau^{(n)} = 0$ and $\tau^{(n)} = 0.4$. However, by mitigating data scarcity within a cluster, inter-cluster learning improves the initial convergence rate. To benefit from an improved initial convergence rate and avoid steady-state performance degradation, it is possible to reduce the inter-cluster learning parameter progressively. Doing so, the PGFL algorithm with time-varying $\tau^{(n)} = 0.4 \times 0.98^{n}$ has the same initial convergence rate as the PGFL algorithm with fixed $\tau = 0.4$ and attains near-identical steady-state NMSD as the PGFL algorithm with fixed $\tau = 0$.

% Privacy trade-off
Finally, we study the impact of privacy protection on the steady-state NMSD of the PGFL algorithm. Fig. \ref{PGFL_additional_c} shows the NMSD after $200$ iterations versus the initial value of the privacy parameter $\phi_0$. Note that, as seen in Theorem III, a lower value of $\phi_0$ ensures more privacy. We observe that for smaller values of $\phi_0$, the steady-state NMSE of the PGFL algorithm is higher. In fact, a lower total privacy loss bound leads to higher perturbation noise variance and diminishes the learning performance of the algorithm.

%%%%%%%%%%%%%%%%%%%%%%%%%%%%%%%%%%%%%%%%
%\subsection{Simulation Setup for Classification}
\subsection{Experiments for Classification on the MNIST Dataset}

The following experiments were conducted on the MNIST handwritten digits dataset \cite{MNIST}. In those experiments, the learning tasks of the clients associated with different clusters share the same data but have different, related, objective functions. The structure of the server network, as well as the number of clients per server, are identical to the experiments for regression. In the following experiments, the clients of a given cluster use the ADMM for logistic regression to differentiate between two classes. The loss function for the logistic regression is given by
\begin{align}
    \log[\ell_k({\bf X}_k, {\bf y}_k; \wbf_k)] = \frac{-1}{D_k} &\sum_{i=1}^{D_k} \Bigl( y_{k,i} \log[y_{k,i}'] \notag \\
    &+ (1 - y_{k,i}) \log[1 - y_{k,i}'] \Bigr),
\end{align}
with
\begin{align}
    y_{k,i}' = \frac{1}{1 + \exp(- \wbf_k^\trp {\bf x}_{k,i})}.
\end{align}
%To evaluate the performance of the algorithm, we display the test accuracy versus the iteration index.
%\begin{align}
%    \label{Accuracy}
%    acc^{(n)} = \sum_{k \in \Cmc} \frac{D_k - m_k^{(n)}}{D_k},
%\end{align}
%where $m_k$ is the number of samples misclassified by the model $\wbf_k$ in the dataset of client $k$.

%%%%%%%%%%%%%%%%%%%%%%%%%%%%%%%%%%%%%%%%
%\subsection{Experiments for Classification on the MNIST Dataset}

% Can be added: MNIST1 - subtasks
%\begin{figure*}[t!]
%    \centering
%    \subfigure[]{\includegraphics[width=0.32\textwidth]{Figures/Journal/PGFL_MNIST_far.eps}}
%    \subfigure[]{\includegraphics[width=0.32\textwidth]{Figures/Journal/PGFL_MNIST_close.eps}}
%    \subfigure[]{\includegraphics[width=0.32\textwidth]{Figures/Journal/PGFL_MNIST_Taus.eps}}
%    \caption{Performance of the PGFL algorithm in the MNIST classification task. Learning curves with and without inter-cluster (a) with low task similarity, (b) with high task similarity, and (c) accuracy after $100$ iterations as a function of $\tau$. }
%    \label{PGFL_MNIST}
%\end{figure*}

\begin{figure}
    \centering
    \includegraphics[width=0.35\textwidth]{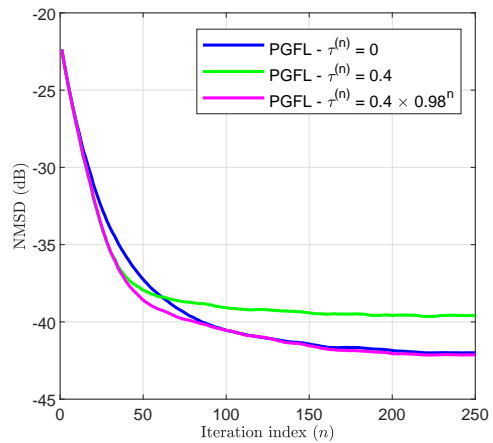}
    \caption{Learning curves of the PGFL algorithm with fixed and time-varying inter-cluster learning parameter $\tau^{(n)}$ in a setting with low cluster similarity, considering client scheduling and privacy.}
    \label{PGFL_additional_b}
\end{figure}
\begin{figure}
    \centering
    \includegraphics[width=0.35\textwidth]{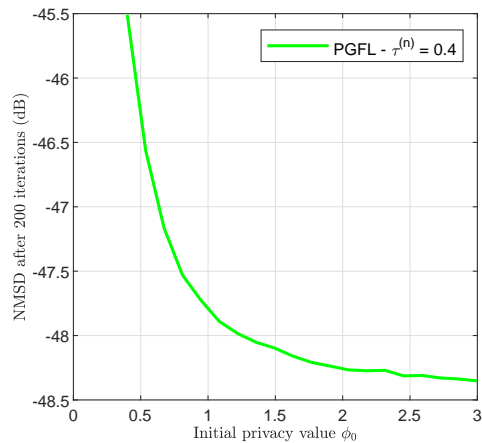}
    \caption{Privacy-accuracy trade-off of the PGFL algorithm with a fixed inter-cluster learning parameter, considering client scheduling.}
    \label{PGFL_additional_c}
\end{figure}
\begin{figure}
    \centering
    \includegraphics[width=0.35\textwidth]{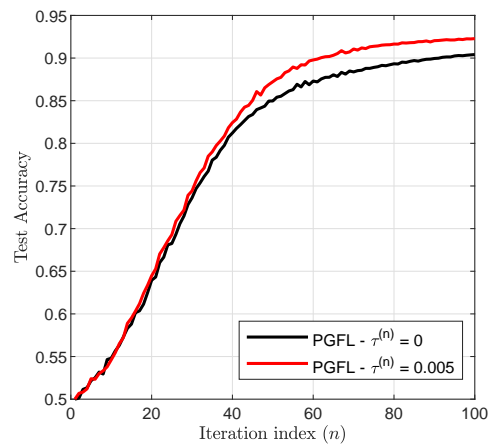}
    \caption{Test accuracy curve of the PGFL algorithm with a fixed inter-cluster learning parameter, considering client scheduling and privacy, with low cluster similarity.}
    \label{PGFL_MNIST_a}
\end{figure}

% MNIST 3, similar tasks 
We simulated the PGFL algorithm in the context of classification with client scheduling, privacy, a fixed inter-cluster learning parameter $\tau^{(n)} = \tau = 0.4$, and without inter-cluster learning $\tau^{(n)} = 0$. Figure \ref{PGFL_MNIST_a} shows the test accuracy versus iteration index in a setting the clients of a given cluster must differentiate between two classes composed of a single digit. Each client receives between $D_k = 2$ and $D_k = 4$ data samples composed of two MNIST images. The clients of cluster $1$ have access to images of the digits $\{1\}$ and $\{8\}$. The clients of clusters $2$ and $3$ have access to images of the digits $\{1\}$ and $\{9\}$, and $\{7\}$ and $\{8\}$, respectively. Given that the clients of different clusters must differentiate between different digits, the similarity between the learning task is limited. Nevertheless, we observe that inter-cluster learning does improve the accuracy of the PGFL algorithm in this setting. %It does so by leveraging the limited similarity between the learning tasks to alleviate data scarcity.

% MNIST 2, very similar tasks
Further, we modified the setting so that the clusters exhibit more similarity. Figure \ref{PGFL_MNIST_b} shows the test accuracy versus iteration index in a setting where the clients of a given cluster must differentiate between two classes composed of triplets of digits. Each client receives between $D_k = 6$ and $D_k = 12$ data samples, each composed of two triplets of MNIST images. The clients of cluster $1$ must differentiate between the classes $\{1,2,3\}$ and $\{6,7,8\}$, the clients of cluster $2$ between $\{1,2,3\}$ and $\{7,8,9\}$, and the clients of cluster $3$ between $\{1,2,3\}$ and $\{6,8,9\}$. We observe that, in this setting, inter-cluster learning significantly improves the accuracy of the PGFL algorithm.

% MNIST 2 taus
Finally, we utilize the previous setting and evaluate the impact of the value of the inter-cluster learning parameter $\tau^{(n)}$ on the accuracy achieved by the PGFL algorithm in the context of classification. Figure \ref{PGFL_MNIST_c} displays the accuracy achieved by the PGFL algorithm after $100$ iterations versus the value of the inter-cluster learning parameter in the context of the classification task of Fig. \ref{PGFL_MNIST_b}. We observe that, in this setting where the similarity among the learning tasks is high, medium and large fixed values for $\tau^{(n)}$ lead to significant accuracy improvement. However, very large values lead to performance degradation, similar to Fig. \ref{PGFL_additional_a}.

%%%%%%%%%%%%%%%%%%%%%%%%%%%%%%%%%%%%%%%%%%%%%%%%%%%%%%%%%%%%%%%%%%%%%%%%%%%%%%%%
\section{Conclusions}
This paper proposed a framework for personalized graph federated learning in which distributed servers collaborate with each other and their respective clients to learn cluster-specific personalized models. The proposed framework leverages the similarities among clusters to improve learning speed and alleviate data scarcity. Further, this framework is implemented with the ADMM as a local learning process and with local zero-concentrated differential privacy to protect the participants' data from eavesdroppers. Our mathematical analysis showed that this algorithm converges to the exact optimal solution for each cluster in linear time and that utilizing inter-cluster learning leads to an alternative output whose distance to the original solution is bounded by a value that can be adjusted with the inter-cluster learning parameter sequence. Finally, numerical simulations showed that the proposed method is capable of leveraging the graph federated architecture and the similarity between the clusters learning tasks to improve learning performance.

\begin{figure}
    \centering
    \includegraphics[width=0.35\textwidth]{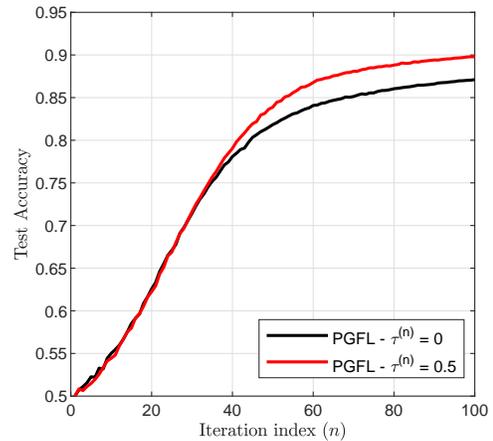}
    \caption{Test accuracy curve of the PGFL algorithm with a fixed inter-cluster learning parameter, considering client scheduling and privacy, with high cluster similarity.}
    \label{PGFL_MNIST_b}
\end{figure}
\begin{figure}
    \centering
    \includegraphics[width=0.35\textwidth]{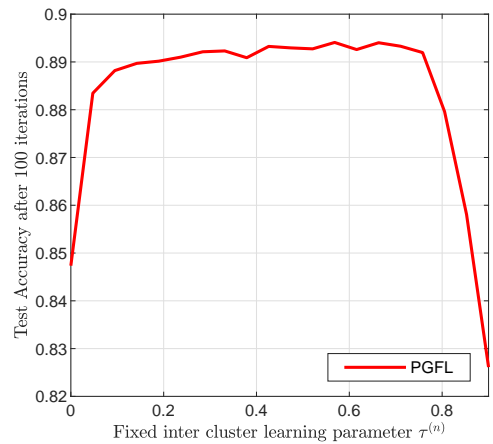}
    \caption{Test accuracy of the PGFL algorithm after $100$ iterations vs. fixed inter-cluster learning parameter $\tau^{(n)}$, considering client scheduling and privacy..}
    \label{PGFL_MNIST_c}
\end{figure}

\bibliographystyle{IEEEtran}
\bibliography{PGFL}

\end{document}